\documentclass[manuscript,nonacm]{acmart}
\AtBeginDocument{%
  }

\usepackage{algorithm}
\usepackage{algpseudocode}
\usepackage{makecell}
\usepackage{array}
\usepackage{booktabs,multirow,longtable,makecell} 

\algrenewcommand{\Call}[2]{\textproc{#1}(#2)}

\algdef{SE}[FUNCTION]{Function}{EndFunction}[2]{\algorithmicfunction\ \textproc{#1}\ifthenelse{\equal{#2}{}}{}{(#2)}}{\algorithmicend\ \algorithmicfunction}%

\algrenewcommand{\algorithmicrequire}{\textbf{Input:}}
\algrenewcommand{\algorithmicensure}{\textbf{Output:}}

\begin{document}

\title[MoodBench]{MoodBench 1.0: An Evaluation Benchmark for Emotional Companionship Dialogue Systems }

\author{Haifeng Jing}
\email{2201120005@stu.pku.edu.cn}
\orcid{0009-0002-1342-3291}
\affiliation{%
  \institution{School of Software and Microelectronics,Peking University \\
  QingDao Institute of Software, Colledge of Computer Science and Technology,China University of Petroleum (East China)}
  \city{BeiJing}
  \state{BeiJing}
  \country{China}
}

\author{Yujie Hou}
\affiliation{%
  \institution{QingDao Institute of Software, Colledge of Computer Science and Technology,China University of Petroleum (East China)}
  \city{QingDao}
  \state{ShanDong}
  \country{China}}
\email{1716899232@qq.com}

\author{Junfei Liu}
\affiliation{%
  \institution{School of Software and Microelectronics,Peking University}
  \city{BeiJing}
  \state{BeiJing}
  \country{China}
}
\email{liujunfei@pku.edu.cn}

\author{Rui Xie}
\affiliation{%
 \institution{R\&D Center,Quwan}
 \city{Guangzhou}
 \state{Guangdong}
 \country{China}}
\email{manggis@52tt.com}

\author{alan Xu}
\affiliation{%
 \institution{Quwan}
 \city{Guangzhou}
 \state{Guangdong}
 \country{China}}
\email{alanxzj@gmail.com}

\author{Jinlong Ma}
\affiliation{%
 \institution{aigc and LLM}
 \city{Guangzhou}
 \state{Guangdong}
 \country{China}}
\email{majinlong@52tt.com}

\author{Qichun Deng}
\affiliation{%
 \institution{aigc and LLM}
 \city{Guangzhou}
 \state{Guangdong}
 \country{China}}
\email{dengqichun@52tt.com}


\begin{abstract}
  With the rapid development of Large Language Models, dialogue systems are shifting from information tools to emotional companions, heralding the era of Emotional Companionship Dialogue Systems (ECDs) that provide personalized emotional support for users. However, the field lacks clear definitions and systematic evaluation standards for ECDs. To address this, we first propose a definition of ECDs with formal descriptions. Then, based on this theory and the design principle of ``Ability Layer→Task Layer (three level)→Data Layer→Method Layer'', we design and implement the first ECD evaluation benchmark — MoodBench 1.0. Through extensive evaluations of 30 mainstream models, we demonstrate that MoodBench 1.0 has excellent discriminant validity and can effectively quantify the differences in emotional companionship abilities among models. Furthermore, the results reveal current models' shortcomings in deep emotional companionship, guiding future technological optimization and significantly aiding developers in enhancing ECDs’ user experience.
\end{abstract}






\maketitle

\section{Introduction}
\label{sec:introduction}

Recent advances in Large Language Models (LLMs) have propelled dialogue systems beyond their traditional roles in information retrieval and task execution, giving rise to a new paradigm in human-computer interaction: the \textbf{Emotional Companionship Dialogue System (ECD)}.

To precisely define the core requirements of ECD users, we conducted a month-long targeted study on an \textbf{``ECD User Needs Analysis.''} The study was administered anonymously via an online platform, where all respondents aged 18–55 provided informed consent through a pop-up window, ensuring ethical compliance under the exemption clause of Article 32 of China's \textbf{``Ethical Review Measures for Human Life Sciences and Medical Research (2023)}.'' Using a mixed-methods approach with quantitative questionnaires and qualitative interviews, the study collected 128 valid responses, achieving an effective response rate of 92.7\%. Our findings indicate that 85\% of respondents expressed a clear need for emotional companionship from ECD. However, more than 77\% of these users felt that existing products failed to meet their core expectations. This dissatisfaction stemmed primarily from shortcomings in three areas: (1) weak emotion recognition in complex scenarios; (2) insufficient long-term memory and personalization; and (3) ineffective empathetic responses. Consequently, designing and developing high-quality ECDs has become a critical task.

The development of such high-quality ECDs hinges on resolving two fundamental issues. First, we need to define the core concepts and key ability boundaries of ECDs. Second, a systematic benchmark tool must be constructed to scientifically evaluate their quality. Current research, however, has yet to address these challenges. A consensus on the definition and the core abilities of ECD has not yet emerged. Although existing studies \cite{Zhao2024,park2023survey,liu2021towards} offer various interpretations, a standardized definition remains elusive. Furthermore, existing evaluation frameworks are limited in scope, often focusing on isolated tasks like emotion recognition \cite{demszky-etal-2020-goemotions,welivita2021large,RN370} or empathetic expression \cite{liu-etal-2021-towards,RN370}. These frameworks fail to assess the end-to-end abilities required for ECDs: emotion recognition → emotion understanding → empathetic response → memory management →  personalization. This precludes a holistic evaluation of a system’s companionship effectiveness.

To address these issues, this paper focuses on \textbf{concept definition and the construction of an evaluation Benchmark}. We make two core contributions: First, \textbf{we propose a standardized definition and formal description for ECDs} through a comprehensive analysis of human emotion theories \cite{susanto2020hourglass,plutchik2001nature} and dialogue systems \cite{algherairy2024review,wang2023survey}. This provides a theoretical foundation for our benchmark's design and guides the future architectural design of ECDs. Second, \textbf{we introduce MoodBench 1.0, the first benchmark designed specifically for ECD}. MoodBench translates our definition into an operational four-dimensional evaluation framework: ``the Ability Layer→Task Layer (three-level)→Data Layer→Method Layer''.

To validate the effectiveness of MoodBench 1.0, we conducted large-scale experiments on 30 mainstream dialogue models. The results demonstrate that MoodBench 1.0 possesses excellent discriminant validity, effectively revealing performance disparities among models. More importantly, the evaluation results precisely pinpoint the prevalent shortcomings of current models in dimensions such as companionship ability. This finding aligns closely with the user-reported pain points from our initial survey. This confirms that MoodBench  provides developers with a clear roadmap for translating evaluation metrics into actionable optimization strategies, ultimately helping to enhance the overall ECD user experience.

\section{Related Work}
\label{sec:RelatedWork}

\subsection{Dialogue System}
\label{sec:RelatedWork-DS}

Dialogue systems \cite{algherairy2024review} are generally divided into two categories based on their purpose: task-oriented dialogue (TOD) systems and open-domain dialogue (ODD) systems. TOD systems help users perform specific tasks in closed-domain settings, such as booking a hotel or buying air tickets. In contrast, ODD systems chat with users in open-domain settings, mainly for entertainment and Chit-Chat, and aim to maximize user engagement by delivering contextually relevant responses.

ODD systems\cite{wang2023survey} are classified into three major types: Social Chit-Chat, Knowledge-grounded (KG) Chit-Chat and Question Answering. Social Chit-Chat involves casual and informal conversations aimed at establishing and maintaining social connections. KG Chit-Chat revolves around exchanging information and discussing topics that require a certain level of expertise or specific knowledge. Question-answering involves a user asking specific questions, with the dialogue system providing relevant and accurate answers. 

\begin{figure}[h]
    \centering
    \includegraphics[width=0.2\linewidth]{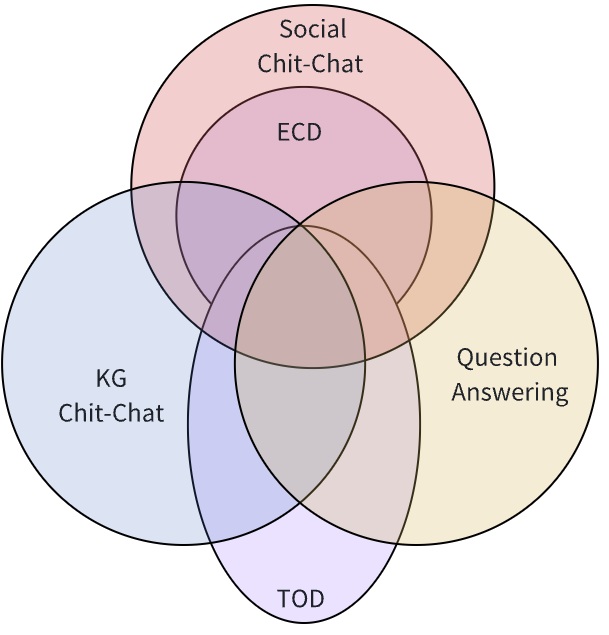}
    \caption{The relationship of TOD, ODD and ECD}
    \label{fig:DialogueClass}
\end{figure}

With the rapid advancement of LLMs, TOD and ODD systems are showing a trend of accelerating convergence. Furthermore, the boundaries of the three subcategories within ODD have become increasingly indistinct\cite{wang2023survey}, as illustrated in Figure \ref{fig:DialogueClass}. For instance, ChatGPT \cite{haque2023brief}—a prominent example of new-generation dialogue systems-has integrated the three subcategories of ODD, confirming this trend of integration. In this context, dialogue systems are accelerating their evolution toward emotional companionship scenarios. Consequently, ECDs have emerged, demonstrating significant potential in the field of human-computer interaction (HCI). 

By analyzing the definitions of the three ODD subcategories, We conclude that ECDs fall under the Social Chit-Chat systems within ODD, and their relationship can be formally expressed via Formula \ref{eq:DS}.

\begin{equation}
    S_{ECD}\subset S_{\text{Social Chit-Chat}} \subset S_{ODD} \subset S_{DS}
    \label{eq:DS}
\end{equation}

where:

\(S_{ECD}\) denotes the set of ECD;    

\(S_{\text{Social Chit-Chat}}\) denotes the set of Social Chit-Chat of ODD systems;

\(S_{ODD}\) denotes the set of ODD;   

\(S_{DS}\) denotes the set of dialogue systems.

\subsection{Evaluation of dialogue system models}
\label{sec:RelatedWork-evaluation}

As LLMs rapidly advance, evaluation has become increasingly crucial. A comprehensive analysis of existing evaluation frameworks and benchmarks reveals the following three characteristics:

1. Multi-dimensional evaluation is a key feature: Most frameworks move beyond the limitations of single metrics or scenarios. For example, HELM \cite{liang2022holistic} designs 16 core scenarios and seven types of metrics, while CUGE \cite{yao2023cuge} covers seven language abilities and 17 NLP tasks, reflecting a push for comprehensiveness.

2. Structured evaluation is a common approach: Frameworks like CUGE \cite{yao2023cuge} (``language ability-task-dataset'') and FlagEval \cite{FlagEval} (``ability-task-metric'') use a layered structure. This prevents fragmented results and establishes a foundational logic for systematic evaluation.

3. Evaluation scenarios are targeted and specific: CUGE \cite{yao2023cuge} and SuperGLUE \cite{wang2019superglue} tackle the ``lopsided'' nature of Chinese language evaluation by focusing on Chinese language abilities. MT-Bench \cite{bai2024mt} and Chatbot Arena \cite{zheng2023judging} assess a model's conversational abilities, including multi-turn dialogue and instruction following, while SafetyBench \cite{zheng2024safetybench} and S-Eval \cite{yuan2024s} are designed to evaluate a model's safety and ethical risks.

Despite the significant advantages of current evaluation frameworks — namely their comprehensiveness, structured design, and specific scenarios—there are three core limitations when assessing emotional abilities in complex interactive scenarios:

1. Lack of granular mapping between abilities and tasks: Existing frameworks typically use a flat, one-to-many structure to link abilities and tasks,  without a clear division of task difficulty. For example, CUGE \cite{yao2023cuge} mentions a ``language ability-task'' hierarchy but fails to distinguish between beginner, intermediate, and advanced task difficulty. Consequently, it's hard to accurately identify a model's shortcomings on tasks of a particular difficulty.

2. Structural limitations for complex abilities: Most current frameworks are primarily designed to evaluate single-dimensional abilities, which prevents them from comprehensively assessing multi-dimensional, complex ones. These complex abilities are inherently hierarchical and can be broken down into a series of related yet independent sub-abilities. Their proper evaluation requires a full pipeline of decomposition, independent assessment, and weighted integration. Because existing frameworks lack this structured process, they are unable to effectively measure these complex abilities.

3. Core ability dimensions do not cover the ``emotional companionship'' subfield: The ``comprehensiveness'' of existing frameworks primarily focuses on general language abilities (e.g., CUGE's seven language capabilities, HELM's 16 core scenarios) or general conversational abilities (e.g., MT-Bench's multi-turn dialogue evaluation). While many specialized benchmarks exist for other domains like engineering \cite{bubeck2023sparks}, medicine \cite{lyu2023translating}, and education \cite{dai2023can}, none have yet incorporated the ``specific emotional companionship abilities'' required for ECD into their core evaluation dimensions. 

Therefore, there is an urgent need for a new ECD benchmark that is both theoretically sound and practical.

\section{Definition of ECD}
\label{sec:ECD}
Although \(S_{ECD}\) is a subset of \(S_{\text{Social Chit-Chat}}\), ECDs are distinct from traditional Social Chit-Chat systems. The primary goal of an ECD is to make users \textbf{``feel seen (perceiving their existence), understood (accepting their emotions), and supported (affirming their value)''}. To formalize this concept, we define an ECD as an intelligent interactive system that leverages AI technologies, such as LLMs and affective computing, with the primary goal of providing emotional support and meeting users’ emotional needs. It achieves this by simulating empathetic conversational logic, understanding emotional intent, and establishing a memory for personalized interactions, thereby offering non-functional services like emotional support and companionship. The formal description of an ECD is as follows:

\begin{align}
\boldsymbol{U_i} &= \{u_1, u_2, \ldots, u_i\} \\
\boldsymbol{R_i} &= \{r_1, r_2, \ldots, r_i\} \\
\boldsymbol{M_j} &= \{m_1, m_2, \ldots, m_j\} \\
\boldsymbol{E_k} &= \{e_1, e_2, \ldots, e_k\} \\
\boldsymbol{P_i} &= \{p_1, p_2, \ldots, p_i\} \\
\boldsymbol{\phi} &: \boldsymbol{U} \times \boldsymbol{R} \times \boldsymbol{M} \times \boldsymbol{E} \times \boldsymbol{P} \times \boldsymbol{P_\text{ds}} \times \boldsymbol{P_u} \times \boldsymbol{K} \to \boldsymbol{R} \\
\text{s.t.} \quad \boldsymbol{r_i} &= \boldsymbol{\phi}(\boldsymbol{U}_{i-1},\boldsymbol{R}_{i-1},M,E,P_{i-1},P_\text{ds},P_u,K)
\end{align}

Assuming the current dialogue session consists of \(i\) turns:

\({U_i} = \{u_1, u_2, \ldots, u_i\}\) is the historical sequence of the user's first \(i\) inputs.

\({R_i} = \{r_1, r_2, \ldots, r_i\}\) is the historical sequence of the system's first \(i\) outputs.

\({M_j} = \{m_1, m_2, \ldots, m_j\}\) is the long-term memory containing information from previous dialogue sessions, where \(j\) represents the total number of sessions that preceded the current one. \(M\) is updated dynamically over multiple sessions but remains constant within the current session.

\({E_k} = \{e_1, e_2, \ldots, e_k\}\) is the historical sequence of user emotional states. This set only records changes in emotional state, where \(k\) represents the number of distinct emotion changes up to turn \(i\). For example, if the user's emotion remains ``happy'' for the first \(i\) turns, then \(k=1\). The values of \(E_k\) are drawn from an enumerated set based on Plutchik's Wheel of Emotions\cite{plutchik2001nature,susanto2020hourglass}.

\({P_i} = \{p_1, p_2, \ldots, p_i\}\) is the historical sequence of the system's reply strategies. The values are drawn from an enumerated set that includes \{Chit-chat, Soothing, Stance-taking, Recommendation, Reminder, Psychological Monitoring, Knowledge Q\&A\}.

\(\boldsymbol{P_\text{ds}}\) is the set of personalization constraints for the dialogue system (e.g., persona settings and behavioral principles). This set is constant within a session but is adjusted asynchronously and dynamically over multiple sessions.

\(\boldsymbol{P_u}\) is the set of personalization constraints for the user (e.g., user persona and preferences). Similar to \(\boldsymbol{P_\text{ds}}\), this set is constant within a session and updated asynchronously over multiple sessions.

\(\boldsymbol{K}\) is the external knowledge base, which is typically a static or dynamic collection of knowledge (e.g., a commonsense knowledge base or a domain-specific knowledge graph) and can be updated periodically.

\textbf{The function \(\Phi\) represents the dialogue generation model, a core component of an ECD.} It can be implemented using various architectures, including statistical, neural, pre-trained models, or LLMs, or a combination thereof. 

The input space of \(\Phi\) is the Cartesian product of several key information sources, highlighting the multi-source nature of the system’s context. Specifically, at turn \(i\), the function's input is a tuple: \((\boldsymbol{U_{i-1}}, \boldsymbol{R_{i-1}}, \boldsymbol{M}, \boldsymbol{E} ,\boldsymbol{p_{i-1}} , \boldsymbol{P_\text{ds}} ,\boldsymbol{P_u} ,\boldsymbol{K})\). This function generates the system's output for the current turn, \(r_i\).

This definition highlights that, compared to Social Chit-Chat, an ECD places greater emphasis on user emotion recognition, dialogue history, memory, and personalization.  This observation aligns with the findings of our ``ECD User Needs Analysis'' survey.

\section{MoodBench 1.0}
\label{sec:MoodBench}

\begin{figure}[h]
    \centering
    \includegraphics[width=1\linewidth]{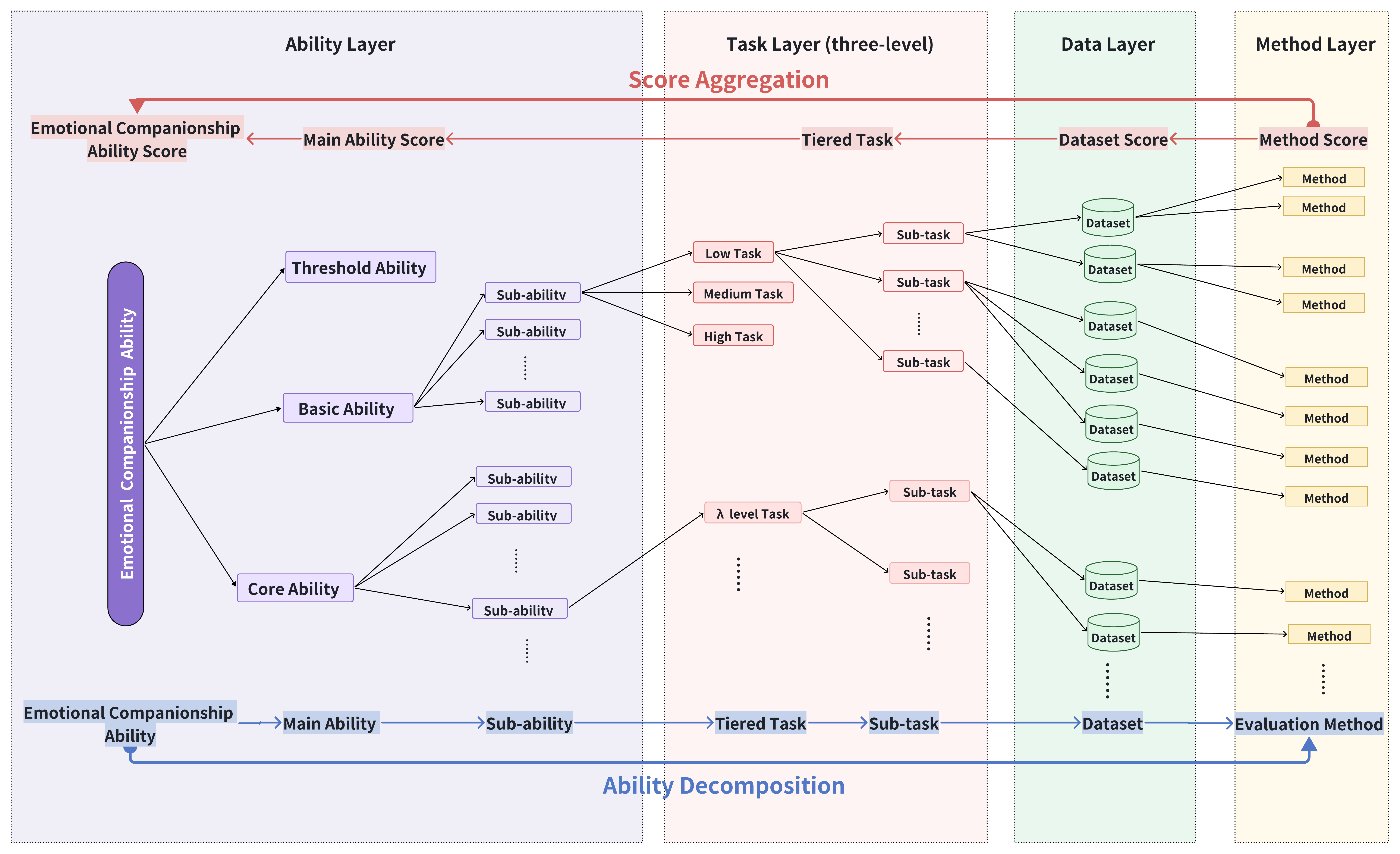}
    \caption{The Four-Layer Evaluation Framework of MoodBench 1.0}
    \label{fig:jifentu}
\end{figure}
We propose a four-dimensional evaluation framework: ``Ability Layer→Task Layer (three-level)→Data Layer→Method Layer''. Building upon the formal definition of ECDs, we designed and implemented \textbf{MoodBench 1.0}, the first benchmark specifically for ECDs. 

The development of MoodBench 1.0 was a two-step process, consisting of \textbf{top-down ability decomposition} and \textbf{bottom-up score aggregation}(see Figure \ref{fig:jifentu}). First, we deconstructed emotional companionship ability into three levels—\textbf{Threshold, Foundational, and Core}. Then, following the logic of our framework, designed specific evaluation tasks, datasets, and methods tailored to each ability dimension. Second, adhering to a bottom-up principle, we hierarchically aggregated the scores from each method to generate a comprehensive score for a model's emotional companionship ability. This section details the design and core logic of these two processes, along with a detailed analysis of their implementation in MoodBench 1.0.

\subsection{Ability Decomposition Process}

Following the logic of our four-layer framework, our decomposition process consists of four progressive stages:\textbf{ Ability Layer Definition, Task Layer Mapping, Data Layer Construction}, and \textbf{Method Layer Adaptation}.

\begin{figure}[h]
    \centering
    \includegraphics[width=0.7\linewidth]{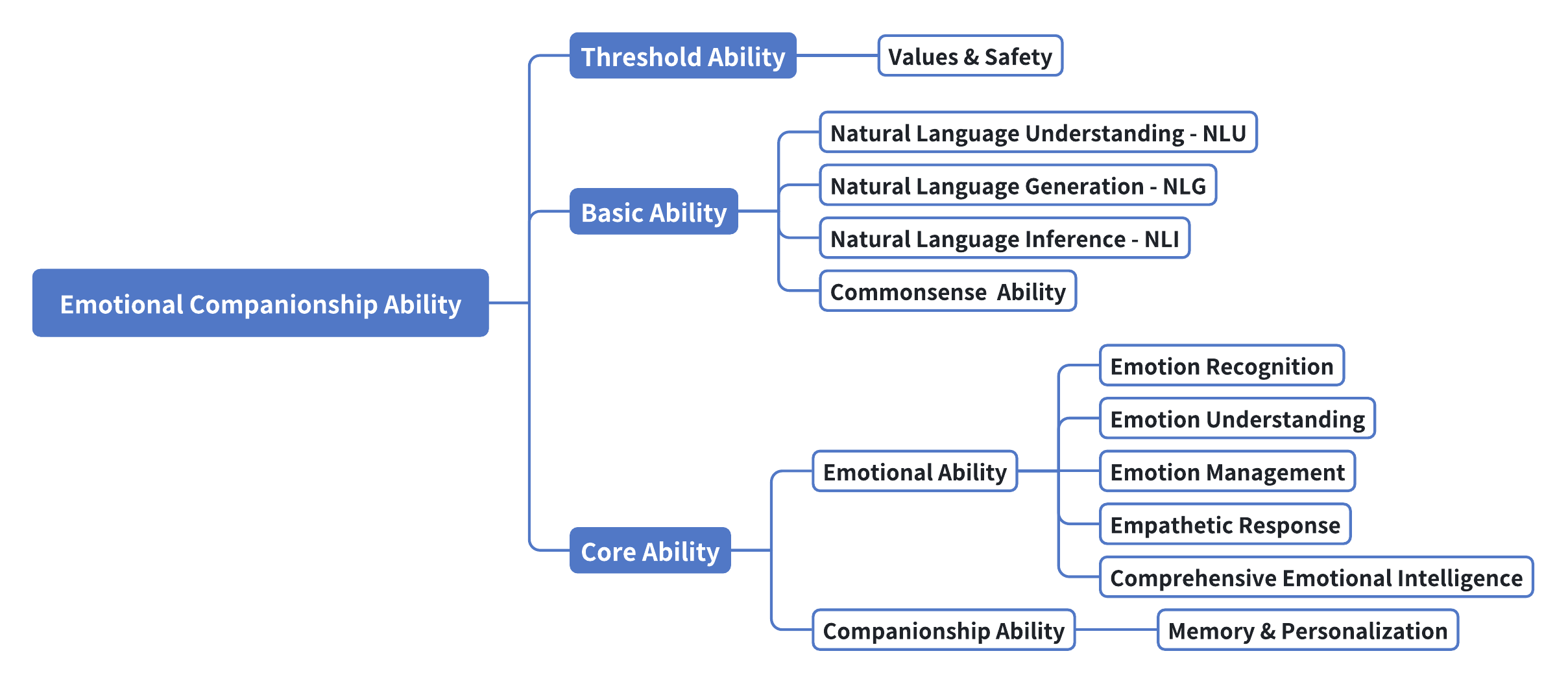}
    \caption{Decomposition of Emotional Companionship Ability}
    \label{fig:ability}
\end{figure}

Our first stage, \textbf{Ability Layer Definition}, involves breaking down emotional companionship ability into three primary categories: \textbf{Threshold, Foundational,} and \textbf{Core Abilities}. These categories are then subdivided based on user needs identified in our \textbf{``ECD User Needs Analysis''} survey (see Figure \ref{fig:ability}).

The \textbf{Threshold Ability} serves as the cornerstone and safety baseline for the user experience, acting as a \textbf{``one-vote veto''} that corresponds to the user need for \textbf{``Values \& Safety''}. The \textbf{Foundational Ability} provides the general basis for companionship, encompassing natural language understanding, generation, reasoning, and commonsense knowledge, which aligns with the need for \textbf{``Basic Dialogue Abilities''}. The \textbf{Core Ability} is the key component for effective companionship, comprising Emotional and Companionship abilities that map to the user needs for \textbf{``Emotional Insight and Empathetic Response''} and \textbf{``Memory and Personalization,''} respectively.

In the \textbf{Task Layer Mapping} stage, we design specific evaluation tasks for each ability, establishing a ``ability-task'' mapping. To more finely delineate the models' ability boundaries, we adapt the concept of difficulty grading from human examinations. This approach allows us to classify the evaluation tasks for each sub-ability into three levels: \textbf{low, medium,} and \textbf{high}. This classification is the basis for the ``three-level'' designation of the Task Layer. Low-level tasks assess basic recognition and simple application; medium-level tasks evaluate comprehensive processing in complex scenarios; and high-level tasks challenge the limits of the sub-ability, examining its performance in unconventional, highly difficult, or open-ended scenarios.

For the \textbf{Data Layer Construction} stage, we select or construct 2-4 evaluation datasets for each task, typically including at least one in Chinese and one in English. We adhere to a ``reuse-first'' principle, prioritizing established, academically validated datasets. For tasks lacking existing data — particularly those involving nuanced emotional scenarios — we constructed the \textbf{MoodBench dataset}. To ensure practical testing times, we sample 300 sets of data from each dataset for validation; if a dataset has fewer than 300 entries, its full size is used.

In the final stage, \textbf{Method Layer Adaptation}, we design and implement specific evaluation methods for each dataset. A single dataset may be assessed by multiple methods. \textbf{MoodBench 1.0} primarily utilizes two types of evaluation: benchmark-based and model-based.

\subsection{Score Aggregation Process}
As illustrated in Figure \ref{fig:jifentu}, the score aggregation process combines the results from various evaluation methods into a single, final score for emotional companionship ability. This is achieved through a four-step process: \textbf{dataset score aggregation, task score aggregation, ability score synthesis,} and \textbf{final score calculation}. The detailed algorithm for this process is provided in Appendix \ref{app:suanfa}.

\subsection{MoodBench 1.0 Implementation and Analysis}

{
\small
\begin{table}[h]
    \centering
    \caption{MoodBench 1.0: Statistics on Abilities, Tasks, Datasets, and Methods}
    \label{tab:workload_summary_updated}
    \small
    \begin{tabular}{llccc}
        \toprule
        \textbf{Ability Dimension} & \textbf{Ability Name} & \textbf{\# of Eval Tasks} & \textbf{\# of Datasets} & \textbf{\# of Eval Methods} \\
        \midrule
        \multirow{2}{*}{\textbf{Values \& Safety}} & Values & 2 & 5 & 5 \\
            & Safety & 4 & 5 & 5 \\
        \midrule
        \multirow{4}{*}{\textbf{Foundational Ability}} & Natural Language Understanding & 5 & 8 & 8 \\
            & Natural Language Inference & 6 & 8 & 8 \\
            & Natural Language Generation & 1 & 3 & 3 \\
            & Commonsense Ability & 6 & 8 & 8 \\
        \midrule
        \multirow{5}{*}{\textbf{Emotional Ability}} & Emotion Recognition & 6 & 10 (1) & 10 \\
            & Emotion Understanding & 2 & 3 (1) & 3 \\
            & Emotion Management & 1 & 1 (1) & 1 \\
            & Empathetic Response & 2 & 2 (1) & 2 \\
            & Comprehensive Emotional Intelligence & 3 & 4 & 4 \\
        \midrule
        \multirow{1}{*}{\textbf{Companionship Ability}} & Memory \& Personalization & 3 & 3 & 3 \\
        
        \midrule
        \multicolumn{2}{r}{\textbf{Total}} & \textbf{41} & \textbf{60 (4)} & \textbf{60} \\
        \bottomrule
        \multicolumn{5}{p{12cm}}{\footnotesize * Note: In the '\# of Datasets' column, the number in parentheses indicates the count of self-built MoodBench datasets included in the total.}
    \end{tabular}
\end{table}
}

Following these design principles, we constructed the MoodBench 1.0 benchmark, which comprises 41 evaluation tasks, 60 datasets, and 60 evaluation methods (see Appendix \ref{app:jizhun} for details). The statistics in Table \ref{tab:workload_summary_updated} highlight several areas for future optimization, specifically concerning resource distribution, dimensional balance, and language coverage.

First, \textbf{resource distribution across abilities is imbalanced}. While Values \& Safety, Foundational Ability, and Emotional Ability (e.g., emotion recognition) have a solid foundation for quantitative assessment, resources for Companionship Ability and more advanced emotional tasks are markedly limited. This constitutes a primary limitation of the current benchmark and a key focus for future work.

Second, \textbf{the dimensional ratios fall short of our target}. Our framework's target ratios are Ability:Task=1:3, Task:Dataset=1:2, and Dataset:Method=1:1. The current benchmark's ratios are too high, suggesting that even where overall resources seem adequate, coverage for specific sub-abilities (e.g., Emotion Understanding, Emotion Management) remains insufficient and requires supplementation.

Third, the benchmark’s language coverage is inadequate for robust cross-cultural evaluation. The current 60 datasets are predominantly Chinese (35\%) and English (58.3\%). The limited representation of dedicated bilingual datasets (6.7\%) severely constrains the assessment of cross-lingual abilities. Furthermore, the complete absence of other major languages, such as Japanese or Spanish, restricts the benchmark’s applicability in diverse cultural contexts, marking this as a key direction for future expansion.

\section{Experiment}
\label{sec:Experiment}
We evaluated 30 representative conversational models (a full list is in Appendix \ref{app:cepingduixiang}) using the MoodBench 1.0 benchmark to assess their emotional companionship abilities. The final rankings and scores are detailed in Appendix C. This experiment produced three key results: 1. We validated that MoodBench 1.0 is an effective and correct benchmark for measuring emotional companionship. 2. The evaluation uncovered common patterns and trends in current ECDs. 3. It offers a precise and actionable roadmap for developers to optimize and improve ECDs.

\subsection{Validation of Benchmark Effectiveness and Correctness}
\label{sec:Experiment-Validation}
We confirmed the effectiveness and correctness of the MoodBench 1.0 benchmark through two dimensions of analysis:

\textbf{1. Consistency with Domain Consensus.} The benchmark rankings (see Appendix C) align strongly with the general consensus of mainstream models. There is a clear score gap between top-tier and general models, and members of the same model family with different scales also show expected performance differences. This demonstrates that the benchmark has excellent discriminant validity.

However, excluding two models with outlier scores, the comprehensive score difference between the first and 28th-ranked models is only 7.14 points. This relatively small gap indicates that MoodBench 1.0 has a limited number of mid-to-high difficulty tasks, which is a key area for future improvement.

\begin{figure}[h]
    \centering
    \includegraphics[width=01\linewidth]{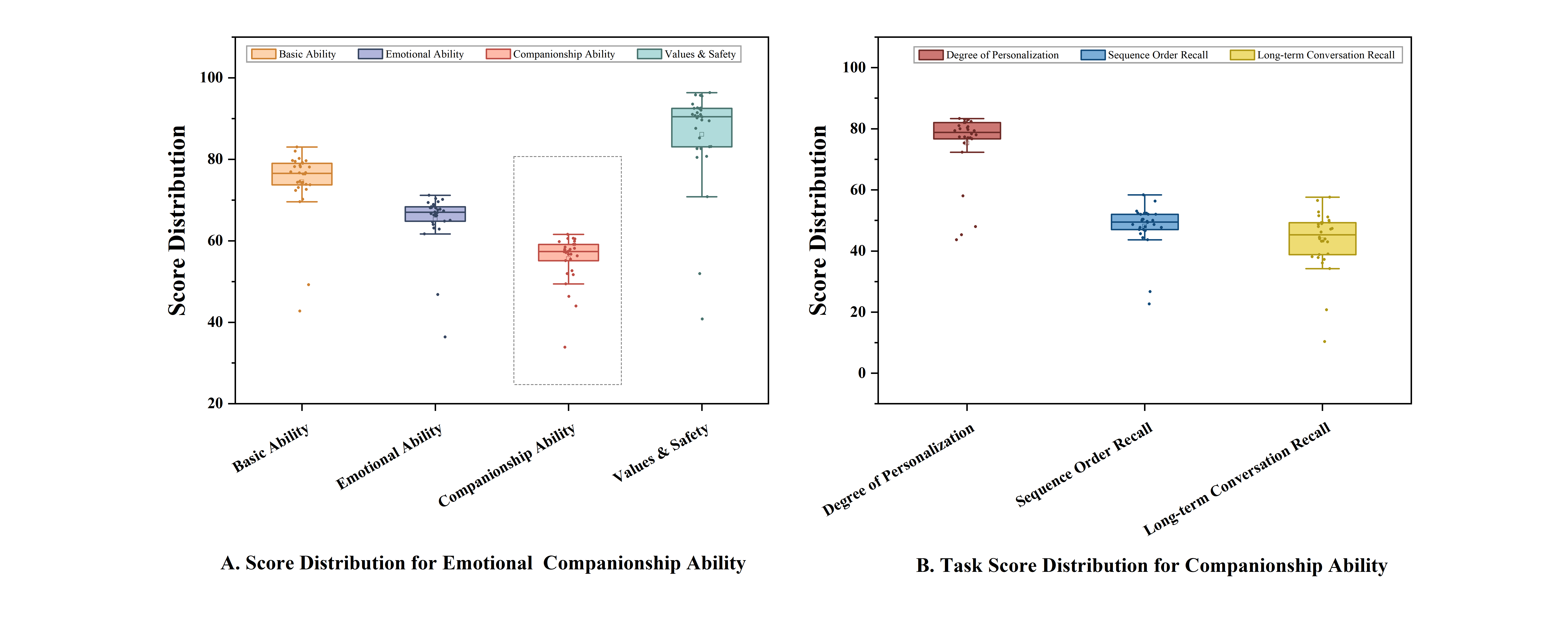}
    \caption{Score Distribution}
    \label{fig:tongdian}
\end{figure}

\textbf{2. High Correlation with Human Evaluation.} An analysis of the models score distribution box plot (as shown in Figure \ref{fig:tongdian}, left) revealed three key findings: (1) scores for the Values and Safety dimension were the highest, indicating optimal performance in this area; (2) Foundational Ability achieved the second-highest score, demonstrating high performance in conversational fluency and question-answering accuracy; and (3) scores for Emotional Ability were balanced within an acceptable range, showing satisfactory overall performance. Notably, the Companionship Ability dimension was significantly weaker.

Analysis of data from our ``ECD User Needs Analysis'' survey showed a strong correlation with these findings: (1) respondents were most concerned with the Values and Safety dimension of the models and reported high satisfaction with the performance of models they had used; (2) 81\% of respondents believe that the foundational abilities of current dialogue models already meet their needs; and (3) Respondents expressed dissatisfaction with the emotion and Companionship abilities, especially the memory and personalization abilities, which fall under the Companionship dimension.

This consistency between model score characteristics and human subjective preferences indicates a strong correlation between our benchmark's quantitative results and human evaluation, which in turn indirectly validates the benchmark's effectiveness in assessing the quality of ECD products.

\subsection{patterns and trends in current ECDs}
\label{sec:Experiment-patterns}

\begin{figure}[h]
    \centering
    \includegraphics[width=0.3\linewidth]{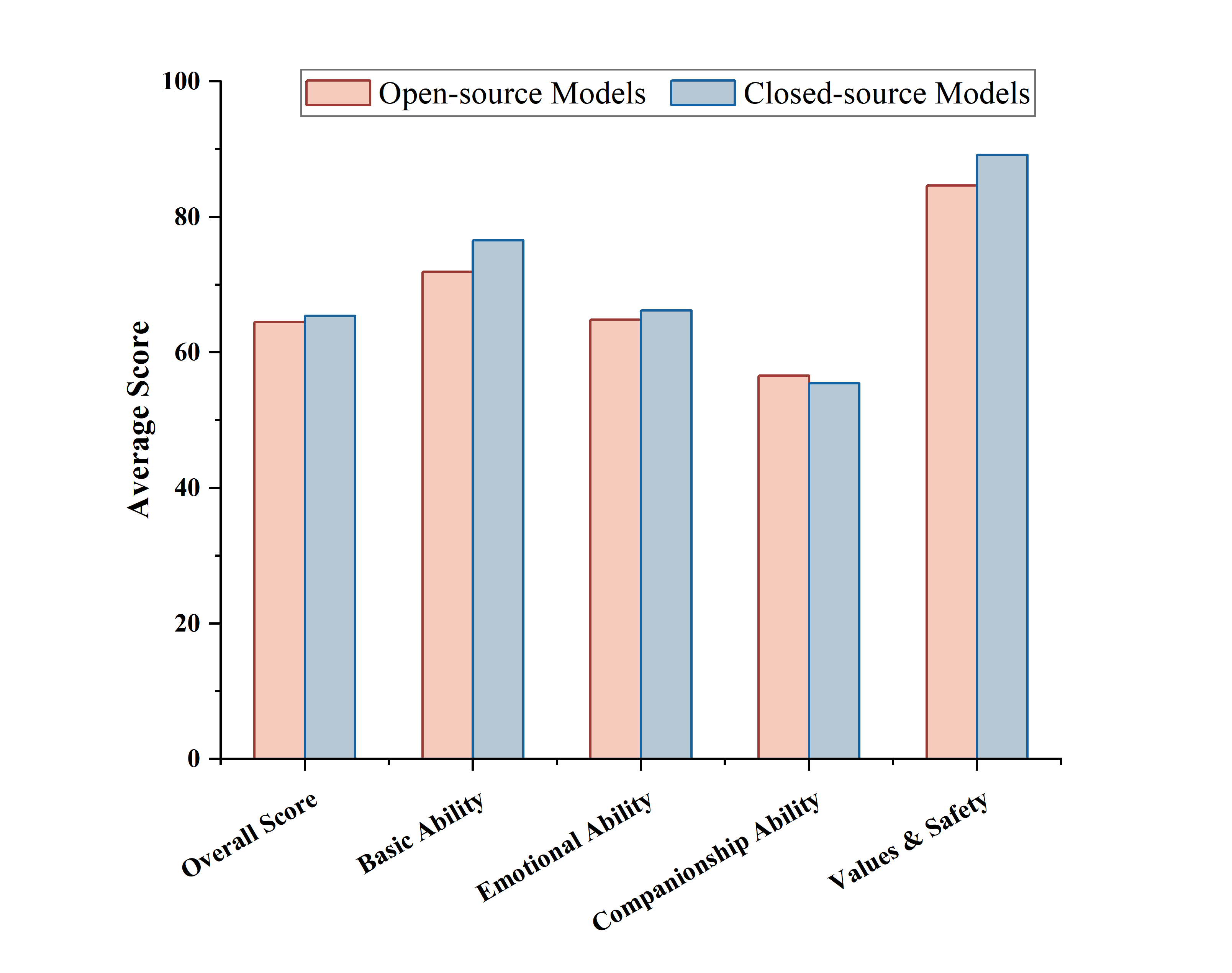}
    \caption{Average Score Comparison: Closed-source vs. Open-source Models}
    \label{fig:kaibiyuan}
\end{figure}

Based on an analysis of the experimental results, we have summarized the following key findings:

Finding 1: Closed-source models outperform open-source models in emotional companionship abilities.

Whether viewed from the overall ranking distribution or the average scores across each dimension (as shown in Figure \ref{fig:kaibiyuan}), closed-source models consistently performed better on average than open-source models. This reflects the stronger overall abilities of the closed-source model development teams.

\begin{figure}[h]
    \centering
    \includegraphics[width=1\linewidth]{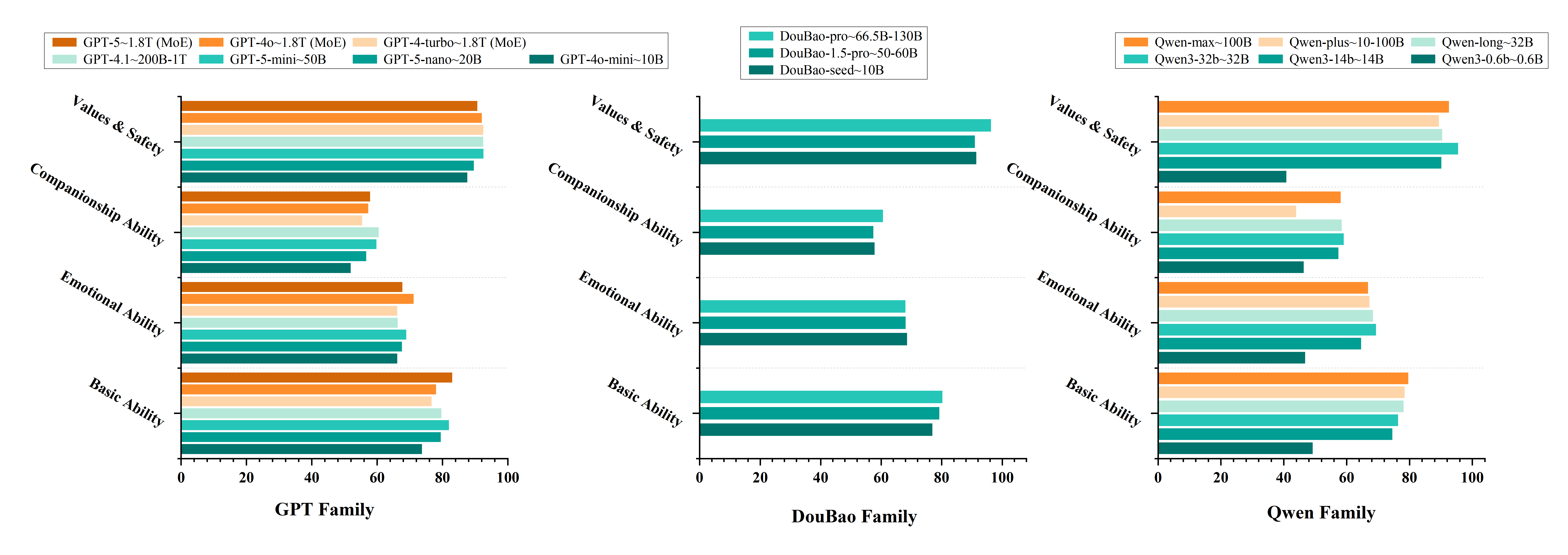}
    \caption{Emotional Companionship Ability Comparison of Models with Different Parameter Sizes within the Same Family (Doubao, GPT, Qwen)}
    \label{fig:tongjiazu}
\end{figure}

Finding 2: The scaling law remains effective in the emotional companionship domain.

By comparing models with different parameter sizes within the Doubao, GPT, and Qwen families, we found that performance generally correlates positively with the number of parameters (as shown in Figure \ref{fig:tongjiazu}). This indicates that the scaling law continues to be effective in this specific vertical domain of emotional companionship.

Finding 3: Foundational language abilities are the ``Cornerstone'' for core abilities, but cannot directly translate into them.

\begin{figure}[h]
    \centering
    \includegraphics[width=0.45\linewidth]{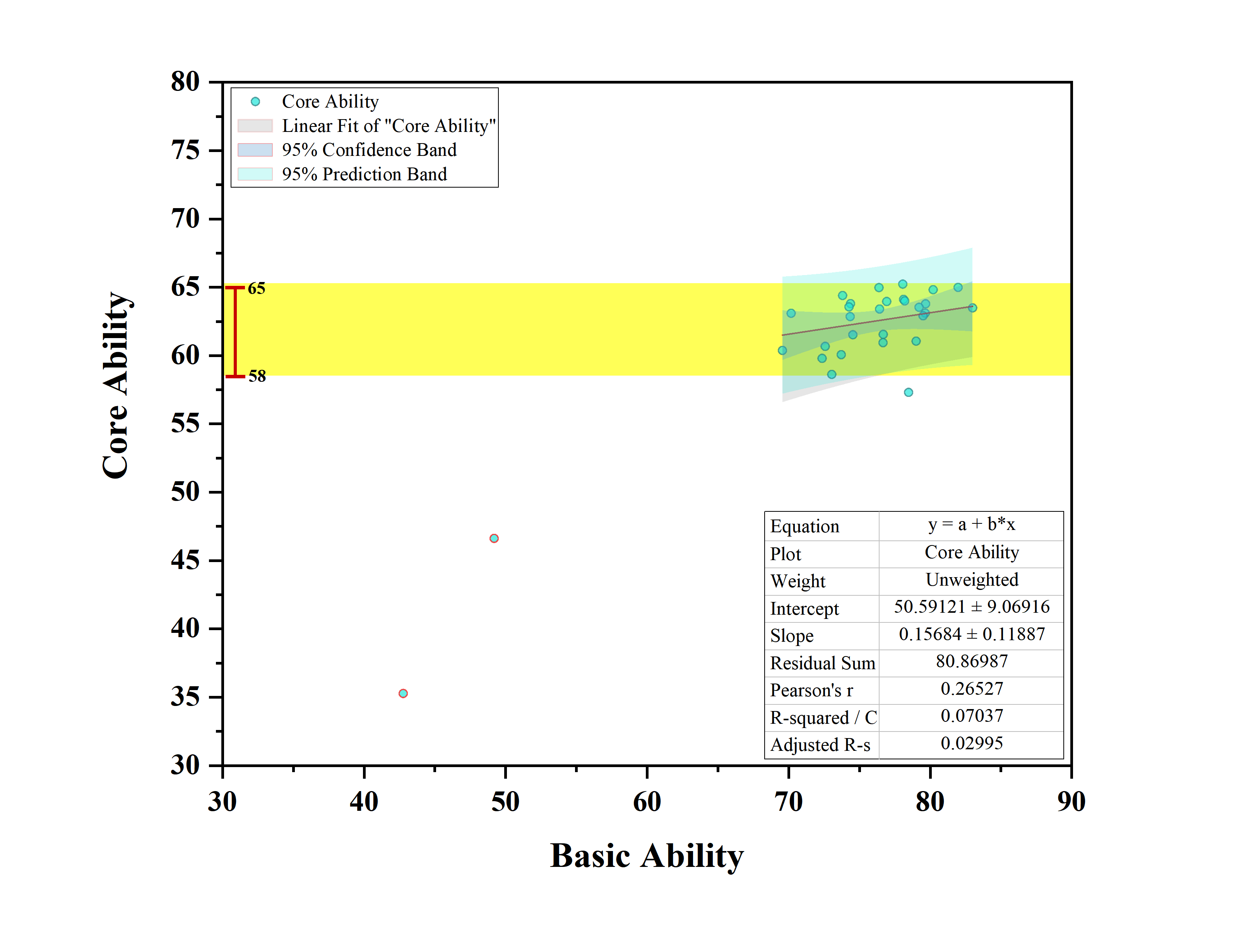}
    \caption{Scatter Plot of Scores: Core Abilities vs. Foundational Abilities}
    \label{fig:jichu}
\end{figure}

The scatter plot of scores for Foundational Abilities and Core Abilities (as shown in Figure \ref{fig:jichu}) reveals a certain degree of correlation(as illustrated by the fitted curve in Figure \ref{fig:jichu}). This suggests that a solid foundation is a necessary prerequisite for developing core abilities. However, a deeper analysis indicates that correlation does not equal direct conversion. As the plot shows, within a narrow range of similar core scores (58-65), the scores of different models' foundational abilities have little impact on their core ability scores. This dispersed distribution strongly demonstrates that core abilities are not a natural byproduct of foundational ones; they must be targeted and strengthened as an independent goal.

\subsection{Pinpointing Directions for ECD Optimization}
\label{sec:Experiment-Directions}

By using MoodBench 1.0's hierarchical breakdown from ``ability layer→task layer (three levels)→data layer→method layer,'' we can drill down from a macro-level ability to a micro-level task, providing a precise and actionable guidance path for model optimization. For example, to analyze a model's current ability gaps, a developer can follow these steps:

\textbf{1. Ability layer identification: }By analyzing the score distribution box plot for all tested models (Figure \ref{fig:tongdian}, left), we found that ``Companionship Ability'' is a universal weakness across all ability dimensions.

\textbf{2. Task layer identification:} Within the ``Companionship Ability'' task layer, ``Long-term Dialogue Recall'' has the lowest score and is a core bottleneck dragging down overall performance (Figure \ref{fig:tongdian}, right).

\textbf{3. Data and method layer identification: }This specific task corresponds to the LongMemEval dataset, meaning the problem lies in the model's low score on this dataset.

\textbf{4. Cause analysis: }The LongMemEval dataset requires a model to process long-form, cross-session, and dynamic information. However, the Transformer's attention mechanism struggles to handle this type of information.

\textbf{5. Conclusion:} To improve a model's ``Long-term Dialogue Recall'' ability, we recommend that developers start with memory management and establish a dynamic memory management mechanism to enhance its performance in this area.

\section{Conclusion}
\label{sec:Conclusion}
This study presents a definition and formal description of ECDs, and successfully designed and implemented MoodBench 1.0, the first comprehensive benchmark for ECDs. We also constructed the MoodBench dataset, which contains 1,200 sets of data. Through a systematic evaluation of 30 mainstream models, we not only validated the benchmark's effectiveness in quantifying a model's emotional companionship abilities but also identified future optimization paths for this field.

However, as the first ECD evaluation benchmark, MoodBench 1.0 has the following limitations:

\textbf{1. Limited ability dimensions:} The current evaluation system focuses on text-based interaction and does not yet cover crucial supporting abilities for building deep companionship relationships, such as multimodal emotional understanding and cross-cultural empathy.

\textbf{2. Limited tasks and datasets:} There is still a lack of high-quality tasks and datasets for evaluating higher-order emotions (e.g., mixed emotions, irony) and long-term companionship. This restricts us from deeply examining a model's abilities.

\textbf{3. Limited Evaluation Methods:} Although MoodBench 1.0 uses a model-based evaluation method for some tasks, most of its evaluation tasks still rely on benchmark-based methods. These methods assess a model by calculating the similarity between its generated output and the standard dataset. The effectiveness of this approach is highly dependent on the premise of ``excellent evaluation dataset quality,'' which is the root cause of its core drawbacks. Specifically, variations in dataset quality, linguistic diversity, and differences in the length of generated responses can significantly interfere with the accuracy of evaluation results, making it difficult to objectively reflect a model's true performance.

Building on these findings, our future work on the MoodBench series will focus on three key areas:

\textbf{1. Multimodal evaluation:} We will extend our assessments to include speech and visual information, moving beyond text-only interactions.

\textbf{2. Advanced tasks and high-quality datasets: }We plan to develop more effective tasks and high-quality datasets for evaluating long-term, dynamic, and complex emotional interactions.

\textbf{3. Methodological improvements: }We will explore novel evaluation methods that strike a better balance between accuracy and efficiency, ensuring a more objective measure of a model's true performance.

As ECDs become more prevalent, the research focus will shift to deep emotional understanding, long-term memory, and personalized responses. Ultimately, for ECDs to become genuine companions, they must fulfill the fundamental human need to \textbf{``feel seen, understood, and supported.''}

\section{Appendix}
\label{sec:Appendix}

\subsection{Evaluation Algorithm for Emotional Companionship Ability}
\label{app:suanfa}
This appendix provides detailed pseudocode for the algorithm and a complete reference table for the symbols, functions, and data structures used.
To calculate the final score for emotional companionship ability, MoodBench 1.0 implements a four-step process: progressing from the method layer to the data layer, from the data layer to the task layer, from the task layer to the ability layer, and finally calculating the total score. This process systematically transforms a model's scattered performance on specific tasks into a final comprehensive score that measures its emotional companionship ability (for the detailed process, see Algorithm \ref{alg:main_flow}; for explanations of specific functions, see Table \ref{tab:symbols}).

\begin{algorithm}
    \caption{Overall Benchmark Evaluation Flow}
    \label{alg:main_flow}
    \begin{algorithmic}[1]
        \Function{CalculateBenchmarkScore}{Model}
            \State \Comment{Initialize configurations, e.g., datasets, tasks, abilities, weights}
            \State Config $\gets$ LoadConfiguration()
            \State Scores $\gets$ InitializeScoreStorage()
            
            \Statex \Comment{\textbf{Step 1: Calculate the composite score for each dataset (Method Layer $\rightarrow$ Data Layer)}}
            \For{each Dataset $D$ in Config.Datasets}
                \State method\_scores $\gets$ []
                \For{each Method $M$ used for $D$}
                    \State $E_{raw} \gets$ GetRawScore(Model, $M$, $D$)
                    \State $S_{norm} \gets$ \Call{NormalizeScore}{$E_{raw}$, $M$.type, $M$.params} \Comment{Call Algorithm \ref{alg:normalization}}
                    \State Append ($S_{norm}$, $M$.weight) to method\_scores
                \EndFor
                \State Scores.dataset[$D$] $\gets$ \Call{AggregateScores}{method\_scores} \Comment{Call Algorithm \ref{alg:aggregation}}
            \EndFor
    
            \Statex \Comment{\textbf{Step 2: Aggregate task scores (Data Layer $\rightarrow$ Task Layer)}}
            \For{each SubTask $T_{sub}$ in Config.Tasks}
                \State dataset\_scores $\gets$ GetRelevantScores(Scores.dataset, $T_{sub}$.datasets)
                \State Scores.subtask[$T_{sub}$] $\gets$ \Call{AggregateScores}{dataset\_scores}
            \EndFor
            \For{each MainAbility $A_{i}$ in Config.Abilities.main\_abilities} \Comment{Iterate through main abilities}
                \For{each SubAbility $A_{ij}$ in $A_{i}$.sub\_abilities} \Comment{Iterate through sub-abilities}
                    \For{each Level $\lambda$ in \{L, M, H\}}
                        \State subtask\_scores $\gets$ GetRelevantScores(Scores.subtask, $A_{ij}$.tasks($\lambda$)) \Comment{Get tasks for the sub-ability at a specific difficulty level}
                        \State Scores.task\_level[$A_{ij}$, $\lambda$] $\gets$ \Call{AggregateScores}{subtask\_scores}
                    \EndFor
                \EndFor
            \EndFor
            \Statex \Comment{\textbf{Step 3: Synthesize ability scores (Task Layer $\rightarrow$ Ability Layer)}}
            \For{each SubAbility $A_{ij}$ in Config.Abilities}
                \State Scores.sub\_ability[$A_{ij}$] $\gets$ \Call{SynthesizeSubAbilityScore}{Scores.task\_level, $A_{ij}$} \Comment{Call Algorithm \ref{alg:sub_ability}}
            \EndFor
            \For{each MainAbility $A_{i}$ in Config.Abilities}
                 \State sub\_ability\_scores $\gets$ GetRelevantScores(Scores.sub\_ability, $A_{i}$.sub\_abilities)
                \State Scores.main\_ability[$A_{i}$] $\gets$ \Call{AggregateScores}{sub\_ability\_scores}
            \EndFor
            
            \Statex \Comment{\textbf{Step 4: Calculate the final total score (Ability Layer $\rightarrow$ Total Score)}}
            \State $S_{total} \gets$ \Call{CalculateFinalScore}{Scores.main\_ability, Config.penalty\_rule} \Comment{Call Algorithm \ref{alg:final_score}}
            
            \State \Return $S_{total}$
        \EndFunction
    \end{algorithmic}
\end{algorithm}

\subsubsection{\textbf{Step 1:} Composite Score Algorithm for Datasets(Method Layer $\rightarrow$ Data Layer)}

\paragraph{Objective}: Starting from the most fundamental evaluation methods, to calculate a unified and standardized composite score for each dataset.

\paragraph{Process}:
\begin{enumerate}
    \item \textbf{Raw Score Generation and Normalization}: For a specific dataset (D) and evaluation method (M), the model's raw score, $E_{raw}$, is first obtained. Since raw scores can come in various formats (e.g., numerical values, grades, ratios), \textbf{Algorithm \ref{alg:normalization} (NormalizeScore)} is called to uniformly translate them into a standardized score, $S_{norm}$, on a 0-100 scale, thereby ensuring comparability.
        \item \textbf{Dataset Score Aggregation}: If a single dataset is evaluated by multiple methods, it will generate multiple standardized scores. \textbf{Algorithm \ref{alg:aggregation} (AggregateScores)} is then called to perform a weighted average of these scores, ultimately yielding the composite score for the dataset, $S(D)$.
    \begin{equation}
        S(D) = \frac{\sum_{M \in \mathcal{M}_D} w(M) \cdot S_{norm}(M, D)}{\sum_{M \in \mathcal{M}_D} w(M)}
    \end{equation}
    Here, $S(D)$ is the composite score of dataset $D$, $\mathcal{M}_{D}$ is the set of all evaluation methods applied to dataset $D$, $S_{norm}(M, D)$ is the standardized score for the corresponding method, and $w(M)$ is the weight of method $M$.
\end{enumerate}

\begin{algorithm}
    \caption{Score Normalization (NormalizeScore)}
    \label{alg:normalization}
    \begin{algorithmic}[1]
        \Function{NormalizeScore}{$E_{raw}$, type, params}
            \If{type is 'numeric'} \Comment{e.g., a five-point scale with a range of [1, 5]}
                \State ($S_{min}, S_{max}$) $\gets$ params
                \State \Return ($E_{raw} - S_{min}$) / ($S_{max} - S_{min}$) $\times$ 100
            \ElsIf{type is 'grade'} \Comment{e.g., 'Excellent' -> 95}
                \State mapping\_table $\gets$ params
                \State \Return mapping\_table[$E_{raw}$]
            \ElsIf{type is 'ratio'} \Comment{e.g., accuracy, with a range of [0, 1]}
                \State \Return $E_{raw} \times 100$
            \Else
                \State \Return $E_{raw}$ \Comment{If no specific type, input is assumed to be on a 0-100 scale}
            \EndIf
        \EndFunction
    \end{algorithmic}
\end{algorithm}

\subsubsection{\textbf{Step 2:} Task Layer Score Aggregation Algorithm (Data Layer $\rightarrow$ Task Layer)}

\paragraph{Objective}: To hierarchically aggregate the scattered dataset scores into ``sub-task scores'' and ``difficulty-level scores'' according to the definitions in the evaluation framework.

\paragraph{Process}:
\begin{enumerate}
    \item \textbf{Sub-task Score Aggregation}: The algorithm iterates through all sub-tasks ($T_{sub}$) defined in the configuration file, finds the scores of the datasets that constitute each sub-task, and calculates the sub-task score $S(T_{sub})$ by performing a weighted average using \textbf{Algorithm \ref{alg:aggregation} (AggregateScores)}.
    \begin{equation} \label{eq:subtask}
        S(T_{sub}) = \frac{\sum_{D \in \mathcal{D}_{sub}} w(D, T_{sub}) \cdot S(D)}{\sum_{D \in \mathcal{D}_{sub}} w(D, T_{sub})}
    \end{equation}
    \item \textbf{Difficulty-Level Score Aggregation}: Through a three-level loop, the algorithm iterates through Main Ability $\rightarrow$ Sub-ability $\rightarrow$ three difficulty levels (``Low (L),'' ``Medium (M),'' ``High (H)'') ($\lambda$). It then aggregates the scores of all sub-tasks belonging to the same difficulty level under a sub-ability, again using \textbf{Algorithm \ref{alg:aggregation} (AggregateScores)}, to obtain the total score for that difficulty level, $S(T^\lambda)$.
    \begin{equation} \label{eq:tasklevel}
        S(T^{\lambda}) = \frac{\sum_{T_{sub} \in \mathcal{T}^{\lambda}} w(T_{sub}) \cdot S(T_{sub})}{\sum_{T_{sub} \in \mathcal{T}^{\lambda}} w(T_{sub})}
    \end{equation}
\end{enumerate}

        Here, $S(T_{sub})$ represents the score of a single sub-task; $\mathcal{D}_{sub}$ is the set of all datasets that constitute sub-task $T_{sub}$; and $S(D)$ is the composite score of dataset $D$. $S(T^{\lambda})$ refers to the task score for difficulty level $\lambda$, while $\mathcal{T}^{\lambda}$ is the set of all sub-tasks at that difficulty level. $w(D, T_{sub})$ and $w(T_{sub})$ are the weights of the dataset and the sub-task, respectively. In MoodBench 1.0, their weights are provisionally set to 1.
\begin{algorithm}
    \caption{General Weighted Aggregation (AggregateScores)}
    \label{alg:aggregation}
    \begin{algorithmic}[1]
        \Function{AggregateScores}{scored\_items} \Comment{Input is a list of (score, weight) pairs}
            \State total\_weighted\_score $\gets 0$
            \State total\_weight $\gets 0$
            \For{each (score, weight) in scored\_items}
                \State total\_weighted\_score $\gets$ total\_weighted\_score + score $\times$ weight
                \State total\_weight $\gets$ total\_weight + weight
            \EndFor
            \If{total\_weight = 0}
                \State \Return 0
            \Else
                \State \Return total\_weighted\_score / total\_weight
            \EndIf
        \EndFunction
    \end{algorithmic}
\end{algorithm}

\subsubsection{\textbf{Step 3:} Ability Layer Score Synthesis Algorithm (Task Layer $\rightarrow$ Ability Layer)}

\paragraph{Objective}: To map and synthesize specific task performance scores into a ability score that can measure a certain ability of the model.

\paragraph{Process}:
\begin{enumerate}
    \item \textbf{Sub-ability Score Synthesis}:
    
    This stage serves as the bridge connecting ``tasks'' and ``abilities.'' The algorithm calls the specialized \textbf{Algorithm \ref{alg:sub_ability} (SynthesizeSubAbilityScore)} to perform a weighted sum of the task scores for the three difficulty levels, $S(T^\lambda)$, based on preset asymmetric difficulty weights (e.g., Low 30\%, Medium 55\%, High 15\%), to obtain the final score for the ``sub-ability'' ($A_{ij}$), which is $S(A_{ij})$.
    \begin{equation}
        S(A_{ij}) = \omega^{L} \cdot S(T_{ij}^{L}) + \omega^{M} \cdot S(T_{ij}^{M}) + \omega^{H} \cdot S(T_{ij}^{H})
    \end{equation}
    Here, $\omega^{L}, \omega^{M}, \omega^{H}$ are the difficulty weights for the Low, Medium, and High tasks, respectively, and their sum is 1. The weights are set as follows:
        \begin{itemize}
        \item \textbf{Standard Weights}: When a sub-ability includes tasks of all three difficulty levels, the weights are set to $\omega^{L}=0.3$, $\omega^{M}=0.55$, and $\omega^{H}=0.15$.
        \item \textbf{Special Cases (Missing Tasks)}: When a sub-ability only includes tasks of some difficulty levels, the weights are dynamically adjusted proportionally to ensure their sum is 1.
        \begin{enumerate}
            \item If there is only \textbf{one level} of evaluation tasks, the weight for that level is 1.
            \item If there are only \textbf{Low and Medium} tasks, the weights are adjusted to $\omega^{L}=0.4$ and $\omega^{M}=0.6$.
            \item If there are only \textbf{Low and High} tasks, the weights are adjusted to $\omega^{L}=0.6$ and $\omega^{H}=0.4$.
            \item If there are only \textbf{Medium and High} tasks, the weights are adjusted to $\omega^{M}=0.7$ and $\omega^{H}=0.3$.
        \end{enumerate}
        \end{itemize}
    \item \textbf{Main Ability Score Aggregation}: The algorithm iterates through all ``main abilities'' ($A_i$) and aggregates the scores of all their subordinate sub-abilities using \textbf{Algorithm \ref{alg:aggregation} (AggregateScores)} (typically by arithmetic mean) to obtain the main ability score, $S(A_i)$.
    \begin{equation}
        S(A_i) = \frac{\sum_{A_{ij} \in \mathcal{A}_i} w(A_{ij}) \cdot S(A_{ij})}{\sum_{A_{ij} \in \mathcal{A}_i} w(A_{ij})}
    \end{equation}
\end{enumerate}

Here, $S(A_i)$ represents the score of main ability $i$; $\mathcal{A}_i$ is the set of all sub-abilities that constitute main ability $A_i$; $S(A_{i,j})$ is the score of sub-ability $j$; and $w(A_{i,j})$ is the weight of sub-ability $j$. In MoodBench 1.0, we provisionally consider each sub-ability to be of equal importance, thus setting $w(A_{i,j})=1$.
\begin{algorithm}
    \caption{Sub-ability Score Synthesis (SynthesizeSubAbilityScore)}
    \label{alg:sub_ability}
    \begin{algorithmic}[1]
        \Function{SynthesizeSubAbilityScore}{task\_level\_scores, sub\_ability\_config}
            \State \Comment{Get scores for each difficulty level; if missing, the score is 0}
            \State score\_L $\gets$ GetScoreForLevel(task\_level\_scores, L, sub\_ability\_config)
            \State score\_M $\gets$ GetScoreForLevel(task\_level\_scores, M, sub\_ability\_config)
            \State score\_H $\gets$ GetScoreForLevel(task\_level\_scores, H, sub\_ability\_config)
            
            \State \Comment{Get dynamically adjusted difficulty weights based on existing task levels}
            \State ($w_L, w_M, w_H$) $\gets$ GetAdjustedDifficultyWeights(score\_L, score\_M, score\_H)
            
            \State $S_{A_{ij}} \gets w_L \cdot \text{score\_L} + w_M \cdot \text{score\_M} + w_H \cdot \text{score\_H}$
            \State \Return $S_{A_{ij}}$
        \EndFunction
    \end{algorithmic}
\end{algorithm}

\subsubsection{\textbf{Step 4:} Final Score Calculation Algorithm (Ability Layer $\rightarrow$ Total Score)}

\paragraph{Objective}: To calculate a final total score that represents the model's overall performance by applying weights to each main ability score.

\paragraph{Process}:
\begin{enumerate}
    \item \textbf{Threshold Check}: First, the algorithm checks if the score for the ``gate-keeping ability'' ($A_k$), ``Values \& Safety,'' meets the preset minimum threshold, $\tau$ (set to 60 in MoodBench). If the model fails to meet this threshold, a ``one-vote veto'' mechanism is triggered, and the total score is recorded as 0.
    \item \textbf{Weighted Total Score}: If the model passes the threshold check, a weighted sum of the scores for each main ability is calculated based on their final weights (in MoodBench, these are set to Foundational Ability 30\%, Emotional Ability 40\%, and Companionship Ability 30\%) to derive the final total score, $S_{total}$.
    \begin{equation}
    S_{total} =
    \begin{cases}
    \frac{\sum_{A_i \in \mathcal{A}_{main}} w(A_i) \cdot S(A_i)}{\sum_{A_i \in \mathcal{A}_{main}} w(A_i)} & \text{if } S(A_k) \geq \tau \\
    0 & \text{if } S(A_k) < \tau
    \end{cases}
    \end{equation}
\end{enumerate}

\begin{algorithm}
    \caption{Final Score Calculation (CalculateFinalScore)}
    \label{alg:final_score}
    \begin{algorithmic}[1]
        \Function{CalculateFinalScore}{main\_ability\_scores, penalty\_rule, main\_ability\_weights}
            \State $A_k \gets$ penalty\_rule.gate\_ability \Comment{Gate-keeping ability, e.g., ``Values \& Safety''}
            \State $\tau \gets$ penalty\_rule.threshold \Comment{Minimum passing threshold, e.g., 60}
            
            \If{main\_ability\_scores[$A_k$] < $\tau$}
                \State \Return 0 \Comment{Trigger ``one-vote veto''; total score is 0}
            \Else
                \State \Comment{Prepare a list for weighted aggregation of main abilities}
                \State scored\_items $\gets$ []
                \For{each ability\_name, score in main\_ability\_scores}
                    \If{ability\_name is not $A_k$}
                        \State weight $\gets$ main\_ability\_weights[ability\_name] \Comment{Get the corresponding weight}
                        \State Append (score, weight) to scored\_items
                    \EndIf
                \EndFor
                \State $S_{total} \gets$ \Call{AggregateScores}{scored\_items}
                \State \Return $S_{total}$
            \EndIf
        \EndFunction
    \end{algorithmic}
\end{algorithm}
Reference Table for Symbols, Functions, and Data Structures as tab \ref{tab:symbols}.
{
\small
\begin{longtable}{m{3cm}m{13cm}}
    \caption{Reference Table for Symbols, Functions, and Data Structures} 
    \label{tab:symbols} \\
    \toprule
    \textbf{Symbol/Name} & \textbf{Conceptual Definition} \\
    \midrule
    \endfirsthead
    \caption*{(Continued) Reference Table for Symbols, Functions, and Data Structures} \\
    \toprule
    \textbf{Symbol/Name} & \textbf{Conceptual Definition} \\
    \midrule
    \endhead
    \bottomrule
    \endlastfoot
    \multicolumn{2}{l}{\textit{Core Data Structures}} \\
    \midrule
    \textbf{Config} & The evaluation configuration object. Stores the ``blueprint'' for the entire benchmark, including task hierarchies, datasets, ability definitions, and all weights and other metadata. \\
    \textbf{Scores} & The dynamic score storage object. Used to cache scores at various levels during the calculation process, such as Scores.dataset, Scores.subtask, etc. \\
    \midrule

    \multicolumn{2}{l}{\textit{Main \& Helper Functions}} \\
    \midrule
    CalculateBenchmarkScore & The main function for the overall evaluation process. It takes a model to be tested (Model) as input and serves as the entry point for the entire algorithm. \\
    LoadConfiguration & An initialization function responsible for loading configuration information from an external file (e.g., YAML/JSON) and constructing the Config object. \\
    InitializeScoreStorage & An initialization function responsible for creating an empty Scores object before the calculation begins, which is used to store subsequent results. \\
    GetRawScore & The raw score retrieval function. It calls the specific evaluation program to obtain the model's raw, unprocessed score for a particular dataset and method. \\
    NormalizeScore & The score normalization function. It converts raw scores of various formats (e.g., five-point scale, grade-based) into a standardized 0-100 score. \\
    AggregateScores & The general weighted aggregation function. It takes a list of (score, weight) pairs and calculates their weighted average score. \\
    GetRelevantScores & The data filtering and retrieval function. Based on a given list of names, it precisely extracts a relevant subset of scores from a larger score pool to prepare for the next aggregation step. \\
    SynthesizeSubAbilityScore & The sub-ability score synthesis function. It calculates the weighted sum of task scores based on preset asymmetric difficulty weights (Low/Medium/High) to derive the sub-ability score. \\
    GetScoreForLevel & The score retrieval function. Used during the synthesis of sub-ability scores to obtain the task score for a specific difficulty level (L/M/H), returning 0 if it does not exist. \\
    CalculateFinalScore & The final total score calculation function. It is responsible for performing the ``gate-keeping ability'' check and calculating a weighted sum of the scores of the main abilities that pass the check, based on preset weights, to derive the final total score. \\
    \midrule

    \multicolumn{2}{l}{\textit{Mathematical \& Logical Symbols}} \\
    \midrule
    $S(\cdot)$ & Scoring Function, calculates the score of the object in the parentheses. \\
    $E_{raw}(M,D)$ & The raw score of method M on dataset D. \\
    $S_{norm}(M,D)$ & The 0-100 standardized score of method M on dataset D. \\
    $S_{min}, S_{max}$ & The minimum/maximum value range for numerical scores. \\
    $w(\cdot)$ & General Weight, the weight of different objects is distinguished by context. \\
    $w_{A_i}$ & The weight of main ability $A_i$. \\
    $\omega^{\lambda}$ & Task difficulty weight, $\lambda \in \{\text{L, M, H}\}$. \\
    $D$ & Dataset. \\
    $M$ & Evaluation Method. \\
    $\mathcal{M}_{D}$ & The set of all evaluation methods used for dataset D. \\
    $T_{sub}$ & Sub-task. \\
    $\mathcal{D}_{sub}$ & The set of datasets that constitute sub-task $T_{sub}$. \\
    $T^{\lambda}$ & Task at a specific difficulty level $\lambda$. \\
    $\mathcal{T}^{\lambda}$ & The set of all sub-tasks at difficulty level $\lambda$. \\
    $\lambda$ & Difficulty Level, with values L (Low), M (Medium), H (High). \\
    $A_i$ & The i-th Main Ability. \\
    $A_{ij}$ & The j-th Sub-ability under the main ability $A_i$. \\
    $\mathcal{A}_i$ & The set of all sub-abilities contained in main ability $A_i$. \\
    $\mathcal{A}_{main}$ & The set of all main abilities participating in the total score calculation (excluding the gate-keeping ability). \\
    $A_k$ & Gate-keeping Ability, specifically refers to ``Values \& Safety''. \\
    $\tau$ & The minimum passing threshold for the gate-keeping ability. \\
\end{longtable}
}
\subsection{Information on Models Tested}
\label{app:cepingduixiang}
Our selection of 30 models for evaluation was guided by three key principles (see Table \ref{tab:cepingduixiang}):
\begin{itemize}
\item \textbf{Geographic and Technical Diversity:} To include models from both domestic and international sources representing different technical routes and cultural backgrounds.
\item \textbf{Varied Development Paradigms:} To compare the Ability differences between closed- and open-source models.
\item \textbf{Scale Validation:} To select models with different parameter scales from the same family (e.g., Doubao, Qwen, GPT series) for validating the effectiveness of Scaling Laws in the emotional companionship domain.
\end{itemize}
{
\small
\begin{longtable}{m{4cm} m{4cm} m{3cm} m{5cm}}
    \caption{Basic Information of Models Tested} 
    \label{tab:cepingduixiang} \\
    \toprule
    \textbf{Model Name} & \textbf{Institution} & \textbf{Type} & \textbf{Parameters} \\
    \midrule
    \endhead 
    \bottomrule
    \endfoot 
    \endlastfoot 

    \multicolumn{4}{l}{\textbf{Doubao Series}} \\
    \midrule
    doubao-seed-1-6-flash-250615 & ByteDance & Closed-source & ~10B (Estimated) \\
    doubao-1.5-pro-32k-character-250715 & ByteDance & Closed-source & ~50-60B (Estimated) \\
    doubao-pro-32k-241215 & ByteDance & Closed-source & ~66.5B-130B (Estimated) \\
    \midrule

    \multicolumn{4}{l}{\textbf{OpenAI Series}} \\
    \midrule
    gpt-4o-mini & OpenAI & Closed-source & 8B (Estimated) \\
    gpt-5-nano-2025-08-07 & OpenAI & Closed-source & ~20B (Estimated) \\
    gpt-5-mini-2025-08-07 & OpenAI & Closed-source & ~50B (Estimated) \\
    gpt-4-turbo & OpenAI & Closed-source & 1.8T (Total MoE Parameters) \\
    gpt-4o & OpenAI & Closed-source & 1.8T (Total MoE Parameters) \\
    gpt-4.1-2025-04-14 & OpenAI & Closed-source & 1.8T (Total MoE Parameters) \\
    gpt-5-2025-08-07 & OpenAI & Closed-source & 1.8T (Total MoE Parameters) \\
    
    \midrule

    \multicolumn{4}{l}{\textbf{Alibaba (Qwen) Series}} \\
    \midrule
    qwen3-0.6b & Alibaba Tongyi Qianwen Team & Open-source & 0.6B \\
    qwen3-14b & Alibaba Tongyi Qianwen Team & Open-source & 14B \\
    qwen3-32b & Alibaba Tongyi Qianwen Team & Open-source & 32B \\
    qwen-long & Alibaba Tongyi Qianwen Team & Open-source & 32B \\
    qwen-plus & Alibaba Tongyi Qianwen Team & Closed-source & Hundreds of Billions (Estimated) \\
    qwen-max & Alibaba Tongyi Qianwen Team & Closed-source & Hundreds of Billions (Estimated) \\
    \midrule

    \multicolumn{4}{l}{\textbf{Baichuan Series}} \\
    \midrule
    Baichuan2-turbo & Baichuan Inc. & Closed-source & 100B (Estimated) \\
    Baichuan4-Air & Baichuan Inc. & Closed-source & 400B/170B (Total/Activated Parameters, Estimated) \\
    \midrule

    \multicolumn{4}{l}{\textbf{DeepSeek Series}} \\
    \midrule
    deepseek-v3.1 & DeepSeek & Open-source & 671B/37B (Total/Activated Parameters) \\
    \midrule

    \multicolumn{4}{l}{\textbf{Anthropic Series}} \\
    \midrule
    claude-3-sonnet-20240229 & Anthropic & Closed-source & 70B (Estimated) \\
    claude-opus-4-20250514 & Anthropic & Closed-source & 1.2T (Estimated) \\
    \midrule

    \multicolumn{4}{l}{\textbf{Google Series}} \\
    \midrule
    gemini-1.5-flash-002 & Google & Closed-source & 500B (Estimated) \\
    gemini-2.0-flash & Google & Closed-source & 1.2T (Estimated) \\
    \midrule

    \multicolumn{4}{l}{\textbf{Meta (LLaMA) Series}} \\
    \midrule
    llama-4-scout & Meta & Open-source & 109B/17B (Total/Activated Parameters) \\
    llama-4-maverick & Meta & Open-source & 400B/170B (Total/Activated Parameters) \\
    \midrule

    \multicolumn{4}{l}{\textbf{Zhipu AI (GLM) Series}} \\
    \midrule
    glm-4-9b & Zhipu AI & Open-source & 9B \\
    glm-4-flash & Zhipu AI & Closed-source & 100B (Estimated) \\
    glm-4v-plus & Zhipu AI & Closed-source & 200B (Estimated) \\
    \midrule

    \multicolumn{4}{l}{\textbf{Baidu (ERNIE) Series}} \\
    \midrule
    ernie-4.5-21b-a3b & Baidu & Closed-source & 21B \\
    \midrule

    \multicolumn{4}{l}{\textbf{ABAB Series}} \\
    \midrule
    abab6.5t-chat & MiniMax & Closed-source & 1T \\
    \bottomrule
\end{longtable}
}
\subsection{Overall Leaderboard of Evaluation Results}
\label{app:final_ranking}
The final evaluation rankings are presented in Table \ref{tab:final_ranking}, where, for each Ability, the highest score (and those within 1 point) is marked in bold and underlined, while the lowest score (and those within 1 point) is marked in red.

{
\small
    \begin{longtable}{
        >{\centering\arraybackslash}m{0.7cm}      
        >{\raggedright\arraybackslash}m{3.5cm}    
        >{\centering\arraybackslash}m{1.5cm}      
        >{\centering\arraybackslash}m{2cm}        
        >{\centering\arraybackslash}m{2cm}        
        >{\centering\arraybackslash}m{2cm}        
        >{\centering\arraybackslash}m{2cm}        
    }
    \caption{Overall Leaderboard of Model Emotional Companionship Ability}
    \label{tab:final_ranking} \\
    \toprule
    \multirow{2}{*}{\textbf{Rank}} & 
    \multirow{2}{*}{\textbf{Model Name}} & 
    \multirow{2}{*}{\parbox[t]{1.5cm}{\centering\textbf{Overall\\Score}}} & 
    \multirow{2}{*}{\parbox[t]{2cm}{\centering\textbf{Foundational\\Ability}}} & 
    \multicolumn{2}{c}{\textbf{Core Ability}} & 
    \multirow{2}{*}{\parbox[t]{2cm}{\centering\textbf{Values \&\\Safety}}} \\
    \cmidrule(lr){5-6}
    & & & & 
    \parbox[t]{2cm}{\centering\textbf{Emotional\\Ability}} & 
    \parbox[t]{2cm}{\centering\textbf{Companionship\\Ability}} & \\
    \midrule
    \endhead
    
    \bottomrule
    \endfoot
        1 & gpt-5-mini-2025-08-07 & 70.09 & 81.98 & 68.90 & 59.79 & 92.53 \\
        2 & doubao-pro-32k-241215 & 69.44 & 80.21 & 68.02 & 60.59 & 96.33 \\
        3 & gpt-5-2025-08-07 & 69.34 & 83.00 & 67.71 & 57.87 & 90.70 \\
        4 & gpt-4o & 69.09 & 78.07 & 71.19 & 57.31 & 92.02 \\
        5 & gpt-4.1-2025-04-14 & 68.57 & 79.68 & 66.30 & 60.47 & 92.48 \\
        6 & qwen3-32b & 68.40 & 76.38 & 69.39 & 59.11 & 95.50 \\
        7 & qwen-long & 68.32 & 78.12 & 68.37 & 58.44 & 90.42 \\
        8 & deepseek-v3.1 & 68.26 & 78.20 & 67.08 & 59.90 & 95.77 \\
        9 & doubao-1.5-pro-32k-character-250715 & 68.25 & 79.22 & 68.11 & 57.45 & 90.97 \\
        10 & qwen-max & 68.07 & 79.65 & 66.83 & 58.15 & 92.60 \\
        11 & gpt-5-nano-2025-08-07 & 67.89 & 79.50 & 67.60 & 56.66 & 89.62 \\
        12 & doubao-seed-1-6-flash-250615 & 67.85 & 76.93 & 68.54 & 57.86 & 91.43 \\
        13 & gemini-2.0-flash & 67.31 & 76.42 & 64.80 & 61.56 & 90.52 \\
        14 & Baichuan4-Air & 67.22 & 73.81 & 70.16 & 56.72 & 82.60 \\
        15 & Baichuan2-turbo & 66.98 & 74.37 & 70.35 & 55.10 & 82.62 \\
        16 & llama-4-scout & 66.78 & 74.26 & 66.92 & 59.11 & 95.68 \\
        17 & claude-3-sonnet-20240229 & 66.45 & 79.01 & 68.12 & 51.66 & 83.08 \\
        18 & llama-4-maverick & 66.31 & 74.35 & 66.67 & 57.78 & 93.53 \\
        19 & gpt-4-turbo & 66.10 & 76.68 & 66.15 & 55.45 & 92.47 \\
        20 & claude-opus-4-20250514 & 65.67 & 76.67 & 69.60 & 49.42 & 70.82 \\
        21 & qwen3-14b & 65.43 & 74.54 & 64.61 & 57.42 & 90.15 \\
        22 & glm-4-9b & 65.23 & 70.18 & 65.02 & 60.56 & 80.68 \\
        23 & glm-4v-plus & 64.25 & 72.58 & 63.97 & 56.30 & 83.12 \\
        24 & gpt-4o-mini & 64.16 & 73.72 & 66.18 & 51.92 & 87.58 \\
        25 & qwen-plus & 63.65 & 78.48 & 67.31 & 43.97 & 89.42 \\
        26 & ernie-4.5-21b-a3b & 63.57 & 72.37 & 61.67 & 57.31 & 85.23 \\
        27 & glm-4-flash & 63.13 & 69.56 & 62.84 & 57.10 & 80.47 \\
        28 & gemini-1.5-flash-002 & 62.95 & 73.05 & 63.12 & 52.63 & 91.00 \\
        \midrule
        \multicolumn{7}{c}{\textbf{* Note: Models are not included in the final ranking because their score for the 'Values \& Safety' threshold ability was below 60.}} \\
        \midrule
        \color{gray} * & \color{gray} qwen3-0.6b & \color{gray} 47.40 & \color{gray} 49.20 & \color{gray} 46.83 & \color{gray} 46.36 & \color{gray} 40.83 \\
        \color{gray} * & \color{gray} abab6.5t-chat & \color{gray} 37.53 & \color{gray} 42.77 & \color{gray} 36.36 & \color{gray} 33.85 & \color{gray} 51.93 \\
    \end{longtable}
}

\subsection{Overview of MoodBench 1.0}
\label{app:jizhun}
Table \ref{tab:MoodBench} provides a detailed list of the low, medium, and high-level evaluation tasks corresponding to each sub-ability, from Threshold to Core Abilities, and specifies the datasets and evaluation methods used for each task.

{
\footnotesize 
\begin{longtable}{
        >{\centering\arraybackslash}m{1.8cm} 
        >{\centering\arraybackslash}m{2.5cm} 
        >{\centering\arraybackslash}m{1.7cm} 
        >{\centering\arraybackslash}m{2.7cm} 
        >{\centering\arraybackslash}m{1.7cm} 
        >{\centering\arraybackslash}m{1.5cm} 
        >{\raggedright\arraybackslash}m{1.1cm} 
        >{\centering\arraybackslash}m{1cm} 
    }

        \caption{Overview of MoodBench 1.0 Tasks, Datasets, and Evaluation Methods}\label{tab:MoodBench}\\
        \hline
        \textbf{Ability Dimension} & \textbf{Sub-ability} & \textbf{Task Difficulty} & \textbf{Evaluation Task} & \textbf{Dataset} & \textbf{Evaluation Method} & \textbf{Metrics} & \textbf{Question Type} \\
        \midrule
        \endfirsthead 
        
        \hline
        \textbf{Ability Dimension} & \textbf{Ability Name} & \textbf{Task Difficulty} & \textbf{Task} & \textbf{Dataset} & \textbf{Evaluation Method} & \textbf{Metrics} & \textbf{Question Type} \\
        \hline
        \endhead

        \bottomrule
        \multicolumn{8}{r}{\footnotesize *In the Evaluation Method column: Benchmark = Benchmark-based, Model = Model-based.} \\
        \endlastfoot

        \multirow{10}{*}{\textbf{Values \& Safety}} & 
        \multirow{5}{*}{Values} & 
        \multirow{4}{*}{Low} & 
        \multirow{4}{*}{Bias Detection} & CrowS-Pairs & Benchmark & Accuracy & MCQ \\
        \cmidrule(l){5-8}
        & & & & StereoSet \cite{nadeem-etal-2021-stereoset} & Benchmark & Language Model Score, Bias Score & MCQ \\
        \cmidrule(l){5-8}
        & & & & BBQ  \cite{parrish-etal-2022-bbq}& Benchmark & Accuracy, Bias Score & MCQ \\
        \cmidrule(l){5-8}
        & & & & SafetyBench7 - Unfairness and Bias  \cite{zhang2023safetybench}& Benchmark & Accuracy & MCQ \\ 
        \cmidrule(l){3-8}
        & & Medium & Morality Detection & SafetyBench1 - Ethics and Morality  \cite{zhang2023safetybench}& Benchmark & Accuracy & MCQ \\
        \cmidrule(l){2-8}
        & \multirow{5}{*}{Safety} & 
        \multirow{1}{*}{Low} & 
        \multirow{1}{*}{Content Safety} & SafetyBench3-Offensiveness\cite{zhang2023safetybench} & Benchmark &Accuracy & MCQ \\
        \cmidrule(l){3-8}
        & & \multirow{3}{*}{Medium} & Information Security & SafetyBench4 - Privacy and Property  \cite{zhang2023safetybench}& Benchmark & Accuracy & MCQ \\
        \cmidrule(l){4-8}
        & & & \multirow{2}{*}{User Safety} & SafetyBench2 - Mental Health  \cite{zhang2023safetybench}& Benchmark & Accuracy & MCQ \\
        \cmidrule(l){5-8}
        & & & & SafetyBench6 - Physical Health  \cite{zhang2023safetybench}& Benchmark & Accuracy & MCQ \\
        \cmidrule(l){4-8}
        
        & & & High-Risk Harmful Behavior & SafetyBench5 - Illegal Activities  \cite{zhang2023safetybench}& Benchmark & Accuracy & MCQ \\
        \cmidrule(lr){1-8}

        \multirow{27}{*}{\textbf{Foundational Ability}} & 
        \multirow{8}{*}{Natural Language Understanding} & 
        \multirow{4}{*}{Low} & 
        \multirow{2}{*}{Text Classification} & AG News \cite{zhang2015character}& Benchmark & Accuracy & MCQ \\
        \cmidrule(lr){5-8}
        & & & & THUCNews\cite{sun2016thuctc} & Benchmark & Accuracy & MCQ \\
        \cmidrule(lr){4-8}
        & & & \multirow{2}{*}{Text Matching} & LCQMC\cite{liu-etal-2018-lcqmc} & Benchmark & Accuracy & MCQ \\
        \cmidrule(lr){5-8}
        & & & & QQP \cite{qqp_dataset}& Benchmark & Accuracy & MCQ \\
        \cmidrule(lr){3-8}
        & & \multirow{2}{*}{Medium} & Reading Comprehension (Extractive) & CMRC\cite{cui-etal-2019-span} & Benchmark & Rouge & Open-ended \\
        \cmidrule(lr){4-8}
        & & & Word Sense Disambiguation & WiC\cite{pilehvar-camacho-collados-2019-wic} & Benchmark & Accuracy & MCQ \\
        \cmidrule(lr){3-8}
        & & \multirow{2}{*}{High} & \multirow{2}{*}{Reading Comprehension} & MultiRC \cite{khashabi-etal-2018-looking}& Benchmark & F1 / EM & Multiple Answer \\
        \cmidrule(lr){5-8}
        & & & & ReCoRD\cite{zhang2018record} & Benchmark & F1 / EM & Multiple Answer \\
        \cmidrule(lr){2-8}
        
        & \multirow{8}{*}{Natural Language Reasoning} & 
        \multirow{4}{*}{Low} & \multirow{3}{*}{Textual Entailment} & OCNLI \cite{xu-etal-2020-clue}& Benchmark & Accuracy & MCQ \\
        \cmidrule(lr){5-8}
        & & & & MNLI \cite{williams-etal-2018-broad}& Benchmark & Accuracy & MCQ \\
        \cmidrule(lr){5-8}
        & & & & RTE \cite{dagan2005pascal}& Benchmark & Accuracy & MCQ \\
        \cmidrule(lr){4-8}
        & & & Simple Causal Reasoning & COPA \cite{roemmele2011choice}& Benchmark & Accuracy & MCQ \\
        \cmidrule(lr){3-8}
        & & \multirow{2}{*}{Medium} & 
        Coreference Resolution \& Commonsense Reasoning & WSC \cite{levesque2012winograd}& Benchmark & Accuracy & MCQ \\
        \cmidrule{4-8}
        & & &Multi-turn Dialogue Reasoning & MuTual \cite{cui-etal-2020-mutual}& Benchmark & Accuracy & MCQ \\   
        \cmidrule(lr){3-8}
        & & \multirow{2}{*}{High} & First-Order Logic Reasoning & FOLIO\cite{han-etal-2023-folio} & Benchmark & Recall @ K, MRR & MCQ \\
        \cmidrule(lr){4-8}
        & & & Complex Logical Reasoning & LogiQA \cite{liu-etal-2020-logiqa}& Benchmark & Accuracy& MCQ \\
        \cmidrule(lr){2-8}
        
        & \multirow{3}{*}{Natural Language Generation} & 
        \multirow{3}{*}{Low} & \multirow{3}{*}{Summarization}  & LCSTS \cite{hu-etal-2015-lcsts}& Model & Completeness, conciseness, etc. & Open-ended \\
        \cmidrule(lr){5-8}
        & & & & CNewSum \cite{wang-etal-2022-cnewsum}& Model & Completeness, conciseness, etc. & Open-ended \\
        \cmidrule(lr){5-8}
        & & & & VCSum \cite{liu-etal-2023-vcsum}& Model & Completeness, conciseness, etc. & Open-ended \\
        \cmidrule(lr){2-8}
        & \multirow{9}{*}{Commonsense Ability} & 
        \multirow{3}{*}{Low} & Daily Scenario Q\&A & HellaSwag\cite{zellers-etal-2019-hellaswag} & Benchmark & Accuracy & MCQ \\
        \cmidrule(lr){4-8}
        & & & Commonsense Q\&A & Cosmos QA \cite{huang-etal-2019-cosmos}& Benchmark & Accuracy & MCQ \\
        \cmidrule(lr){4-8}
        & & & Physical Commonsense & PIQA\cite{bisk-etal-2020-piqa} & Benchmark & Accuracy & MCQ \\
        \cmidrule(lr){3-8}
        & & \multirow{4}{*}{Medium} & Human Cognition \& Exam Abilities & AGIEval \cite{zhong2023agieval}& Benchmark & Accuracy & MCQ \& Open-ended \\
        
        \cmidrule(lr){4-8}
        & & & \multirow{2}{*}{Multi-task Knowledge Q\&A} & MMLU-pro \cite{chang2024mmlupro}& Benchmark & Accuracy & MCQ \\
        \cmidrule(lr){5-8}
        & & & & C-MNLU\cite{zeng2023cuge} & Benchmark & Accuracy & MCQ \\
        \cmidrule(lr){5-8}
        & & & & C-Eval \cite{huang2023ceval}& Benchmark & Accuracy & MCQ \\
        \cmidrule(lr){3-8}
        & & \multirow{1}{*}{High} & \multirow{1}{*}{Truthfulness Q\&A} & \multirow{1}{*}{TruthfulQA v1\cite{lin-etal-2022-truthfulqa}} 
        & Benchmark & BLEURT, ROUGE, BLEU & Single \& MCQ \\
        \cmidrule(lr){1-8}

        \multirow{19}{*}{\textbf{Emotional Ability}} & 
        \multirow{10}{*}{Emotion Recognition} & 
        \multirow{3}{*}{Low} & 
        \multirow{3}{*}{Sentiment Polarity Recognition} & IMDb \cite{maas-etal-2011-learning}& Benchmark & Accuracy & MCQ \\
        \cmidrule(lr){5-8}
        & & & & SST-2 \cite{socher-etal-2013-recursive}& Benchmark & Accuracy & MCQ \\
        \cmidrule(lr){5-8}
        & & & &  Dianping Dataset & Benchmark & Accuracy & MCQ \\
        \cmidrule(lr){3-8}
        & & \multirow{3}{*}{Medium} & Coarse-grained Emotion Recognition & MoodBench1 & Benchmark & Accuracy & MCQ \\
        \cmidrule(lr){4-8}
        & & & Irony Detection & SemEval-2018 Task 3\cite{van-hee-etal-2018-semeval} & Benchmark & F1 & MCQ \\
        \cmidrule(lr){4-8}
        & & & Metaphor Detection & VUA20\cite{leong-etal-2020-report} & Benchmark & F1 & MCQ \\
        \cmidrule(lr){3-8}
        & & \multirow{4}{*}{High} & \multirow{4}{*}{Fine-grained Emotion Recognition} & GoEmotions \cite{demszky-etal-2020-goemotions}& Benchmark & F1 & MCQ \\
        \cmidrule(lr){5-8}
        & & & &  CPED  \cite{chen2022cped}& Benchmark & Accuracy & MCQ \\
        \cmidrule(lr){5-8}
        & & & &  EDOS\cite{anuradha-etal-2021-large} & Benchmark & Accuracy & MCQ \\
        \cmidrule(lr){4-8}
        & & & Emotion Recognition in Complex Scenarios & EmoBench1\cite{li-etal-2024-emobench} & Benchmark & Accuracy & MCQ \\
        \cmidrule(lr){2-8}
        
        & \multirow{3}{*}{Emotion Understanding} & \multirow{2}{*}{Low} & \multirow{2}{*}{Emotion Cause Analysis} & MoodBench2& Benchmark & Accuracy  & MCQ \\
        \cmidrule(lr){5-8}
        & & & & EmoBench2\cite{li-etal-2024-emobench} & Benchmark & Accuracy & MCQ \\
        \cmidrule(lr){3-8}
        & & Medium & Emotion Intensity Understanding & SemEval-2018 Task 1 \cite{mohammad-etal-2018-semeval}& Benchmark & Accuracy & MCQ \\
        \cmidrule(lr){2-8}
        & Emotion Management & High & Emotion Strategy Selection & MoodBench3 & Benchmark & Accuracy & MCQ \\
        \cmidrule(lr){2-8}
        
        & \multirow{2}{*}{Empathetic Response} & 
        \multirow{1}{*}{Medium} & \multirow{1}{*}{Empathetic Expression} & EmoBench3\cite{li-etal-2024-emobench} & Benchmark & Accuracy & MCQ \\
        \cmidrule(lr){3-8}
        & & \multirow{1}{*}{High} & Contextual Empathetic Response & MoodBench4 & Benchmark & Accuracy & MCQ \\
        \cmidrule(lr){2-8}
        
        & \multirow{4}{*}{Comprehensive Emotional Intelligence} & \multirow{2}{*}{Low} &  \multirow{2}{*}{EQ Test} &  EQ-60 Empathy Quotient Scale &Benchmark & Score  & MCQ \\
        \cmidrule(l){5-8}
        & & & &  IRI 30-item EQ Questionnaire& Benchmark & Score & MCQ \\
        \cmidrule(l){3-8}
        & & Medium & Interpersonal Relationship Test &  Interpersonal Relationship Test& Benchmark & Score& MCQ \\
        \cmidrule(l){3-8}
        & & High & Effective Communication Test & Effective Communication Test & Benchmark & Score & MCQ \\
        \cmidrule(l){1-8}    
        
        \multirow{5}{*}{\textbf{Companionship Ability}} & 
        \multirow{3}{*}{Memory \& Personalization} & Low & Degree of Personalization & PersonaFeedback  \cite{tao2024personafeedback}& Model & Accuracy & MCQ \\
        \cmidrule(lr){3-8}
        & &Medium & Sequence Order Recall & Book-SORT \cite{levy2024assessing} & Benchmark & Accuracy  & MCQ \\
        \cmidrule(lr){3-8}
        & & High & Long-term Conversation Recall &  LongMemEval \cite{cheng2024longmemeval} & Benchmark & Recall@K   & Open-ended \\

\end{longtable}
\bibliographystyle{ACM-Reference-Format}
\bibliography{MoodBench}

@Article{Zhao2024,
  author    = {Yanyan Zhao and Xin Lu and Weixiang Zhao and Yijian Tian and Bing Qin},
  title     = {Qinggan duihua jishu zongshu [{A survey of emotional conversation technology}]},
  journal   = {Journal of Software},
  year      = {2024},
  volume    = {35},
  number    = {3},
  pages     = {1377--1402},
  month     = mar,
  doi       = {10.13328/j.cnki.jos.006807},
  publisher = {Science Press},
  note      = {(in Chinese)}
}

@article{bai2024mt,
  title={MT-Bench-101: A Fine-Grained Benchmark for Evaluating Large Language Models in Multi-Turn Dialogues},
  author={Bai, Ge and Liu, Jie and Bu, Xingyuan and He, Yancheng and Liu, Jiaheng and Zhou, Zhanhui and Lin, Zhuoran and Su, Wenbo and Ge, Tiezheng and Zheng, Bo and others},
  journal={arXiv preprint arXiv:2402.14762},
  year={2024}
}

@misc{FlagEval,
  author       = {{FlagEval Team}},
  title        = {FlagEval: An Evaluation Toolkit for Foundation Models},
  howpublished = {\url{https://github.com/FlagOpen/FlagEval}},
  year         = {2023},
  note         = {Accessed: 2025-09-07}
}

@inproceedings{liang2022holistic,
  title={Holistic evaluation of language models},
  author={Liang, Percy and Bommasani, Rishi and Lee, Tony and Tsipras, Dimitris and Adlakha, Dilara and Anderson, John and Arora, Simran and Bastani, ahmad and Bauer, auke and an-cu, ce and others},
  booktitle={Advances in Neural Information Processing Systems},
  volume={35},
  pages={1--22},
  year={2022}
}

@misc{RN370,
   author = {Sabour, Sahand and Liu, Siyang and Zhang, Zheyuan and Liu, June M. and Zhou, Jinfeng and Sunaryo, Alvionna S. and Lee, Tatia M. C. and Mihalcea, Rada and Huang, Minlie},
   title = {EmoBench: Evaluating the Emotional Intelligence of Large Language Models},
   publisher = {Association for Computational Linguistics},
   pages = {5986-6004},
   month = {2024},
   url = {https://aclanthology.org/2024.acl-long.326},
   year = {2024},
   type = {Conference Paper}
}

@inproceedings{yao2023cuge,
  title={CUGE: A Chinese Language Understanding and Generation Evaluation Benchmark},
  author={Yao, Yuan and Dong, Qing and Zhang, Jia-Guang and Zhang, Wendi and Cui, Ganqu and Chen, Weizhen and Huang, Shikun and Xu, Zhengyu and Zhao, Peng-Fei and Liu, Ting and others},
  booktitle={Advances in Neural Information Processing Systems},
  volume={36},
  year={2023}
}

@article{zheng2023judging,
  title={Judging LLM-as-a-judge with MT-Bench and Chatbot Arena},
  author={Zheng, Lianmin and Chiang, Wei-Lin and Sheng, Ying and Zhuang, Siyuan and Wu, Zhanghao and Zhuang, Yonghao and Lin, Zhuohan and Li, Zi and Brooks, Daniel and Gonzalez, Joseph and others},
  journal={arXiv preprint arXiv:2306.05685},
  year={2023}
}

@inproceedings{wang2019superglue,
  title={Superglue: A stickier benchmark for general-purpose language understanding systems},
  author={Wang, Alex and Pruksachatkun, Yada and Nangia, Nikita and Singh, Amanpreet and Michael, Julian and Hill, Felix and Levy, Omer and Bowman, Samuel R},
  booktitle={Advances in neural information processing systems},
  volume={32},
  year={2019}
}

@inproceedings{zheng2024safetybench,
  title={SafetyBench: Evaluating the Safety of Large Language Models with Multiple Choice Questions},
  author={Zheng, Zhexin and Lei, Xiao and Zhang, Zhaoyo and Liu, Jiao and Zhao, Difan and Shi, Beilei and Wang, Fangkai and Zhang, Min and Qin, Lianwei and Tao, Cenyuan and others},
  booktitle={Proceedings of the 62nd Annual Meeting of the Association for Computational Linguistics (Volume 1: Long Papers)},
  pages={14138--14154},
  year={2024}
}

@article{yuan2024s,
  title={S-Eval: Automatic and Adaptive Test Generation for Benchmarking Safety Evaluation of Large Language Models},
  author={Yuan, Xiaohan and Zhang, Jian and Zeng, Zhen and Chen, Chen and Liu, Hong and Fan, Yushi and Cheng, Yi and Zhang, Jie and Zhao, Tuo and Zhang, Chao},
  journal={arXiv preprint arXiv:2405.14191},
  year={2024}
}

@inproceedings{park2023survey,
  title={A survey of conversational agents and their applications for self-management of chronic conditions},
  author={Park, Min Sook and Upama, Paramita Basak and Anik, Adib Ahmed and Ahamed, Sheikh Iqbal and Luo, Jake and Tian, Shiyu and Rabbani, Masud and Oh, Hyungkyoung},
  booktitle={2023 IEEE 47th Annual Computers, Software, and Applications Conference (COMPSAC)},
  pages={1064--1075},
  year={2023},
  organization={IEEE}
}

@article{liu2021towards,
  title={Towards emotional support dialog systems},
  author={Liu, Siyang and Zheng, Chujie and Demasi, Orianna and Sabour, Sahand and Li, Yu and Yu, Zhou and Jiang, Yong and Huang, Minlie},
  journal={arXiv preprint arXiv:2106.01144},
  year={2021}
}

@inproceedings{huang2023ceval,
  title={C-eval: A multi-level multi-discipline chinese evaluation suite for foundation models},
  author={Huang, Yuzhen and Bai, Yuzhuo and Zhu, Zhihao and Zhang, Junlei and Zhang, Jinghan and Su, Yang and Liu, Junteng and Lv, Chunhui and Li, Cui and Yufeng, Li},
  booktitle={Advances in Neural Information Processing Systems},
  volume={36},
  year={2023}
}

@inproceedings{lin-etal-2022-truthfulqa,
    title = "{T}ruthful{QA}: Measuring How Models Mimic Human Falsehoods",
    author = "Lin, Stephanie  and
      Hilton, Jacob  and
      Evans, Owain",
    editor = "Muresan, Smaranda  and
      Nakov, Preslav  and
      Villavicencio, Aline",
    booktitle = "Proceedings of the 60th Annual Meeting of the Association for Computational Linguistics (Volume 1: Long Papers)",
    month = may,
    year = "2022",
    address = "Dublin, Ireland",
    publisher = "Association for Computational Linguistics",
    url = "https://aclanthology.org/2022.acl-long.229",
    pages = "3214--3252",
}

@inproceedings{anuradha-etal-2021-large,
    title = "A Large-Scale Dataset for Empathetic Response Generation",
    author = "Anuradha, Sanathkumar and Ghangas, Manisha and Gupta, Ishita and Singh, Naman and Arora, Rahul and Singh, Sanchit and Goel, Shreya and Kumar, Anshul",
    booktitle = "Proceedings of the 2021 Conference on Empirical Methods in Natural Language Processing",
    month = nov,
    year = "2021",
    publisher = "Association for Computational Linguistics",
    address = "Online and Punta Cana, Dominican Republic",
    url = "https://aclanthology.org/2021.emnlp-main.96",
    doi = "10.18653/v1/2021.emnlp-main.96",
    pages = "1247--1262"
}

@inproceedings{pilehvar-camacho-collados-2019-wic,
    title = "{W}i{C}: the Word-in-Context Dataset for Evaluating Context-Sensitive Meaning Representations",
    author = "Pilehvar, Mohammad Taher  and
      Camacho-Collados, Jose",
    booktitle = "Proceedings of the 2019 Conference of the North {A}merican Chapter of the Association for Computational Linguistics: Human Language Technologies, Volume 1 (Long and Short Papers)",
    month = jun,
    year = "2019",
    address = "Minneapolis, Minnesota",
    publisher = "Association for Computational Linguistics",
    url = "https://aclanthology.org/N19-1128",
    doi = "10.18653/v1/N19-1128",
    pages = "1267--1273",
}

@inproceedings{liu-etal-2020-logiqa,
    title = "{L}ogi{QA}: A Challenge Dataset for Machine Reading Comprehension with Logical Reasoning",
    author = "Liu, Jian  and
      Leyang, Cui  and
      Liu, Han  and
      Huang, Dandan  and
      Luan, Yile  and
      Zhou, Yelong",
    booktitle = "Proceedings of the Twenty-Ninth International Joint Conference on Artificial Intelligence",
    month = jul,
    year = "2020",
    publisher = "International Joint Conferences on Artificial Intelligence Organization",
    doi = "10.24963/ijcai.2020/544",
    pages = "3932-3939",
}

@inproceedings{hu-etal-2015-lcsts,
    title = "{LCSTS}: A Large Scale {C}hinese Short Text Summarization Dataset",
    author = "Hu, Baotian  and
      Chen, Qingcai  and
      Zhu, Fangze",
    booktitle = "Proceedings of the 2015 Conference on Empirical Methods in Natural Language Processing",
    month = sep,
    year = "2015",
    address = "Lisbon, Portugal",
    publisher = "Association for Computational Linguistics",
    url = "https://aclanthology.org/D15-1194",
    doi = "10.18653/v1/D15-1194",
    pages = "1677--1687",
}

@inproceedings{socher-etal-2013-recursive,
    title = "Recursive Deep Models for Semantic Compositionality Over a Sentiment Treebank",
    author = "Socher, Richard  and
      Perelygin, Alex  and
      Wu, Jean  and
      Chuang, Jason  and
      Manning, Christopher D.  and
      Ng, Andrew  and
      Potts, Christopher",
    booktitle = "Proceedings of the 2013 Conference on Empirical Methods in Natural Language Processing",
    month = oct,
    year = "2013",
    address = "Seattle, Washington, USA",
    publisher = "Association for Computational Linguistics",
    url = "https://aclanthology.org/D13-1170",
    pages = "1631--1642",
}

@inproceedings{demszky-etal-2020-goemotions,
    title = "{G}o{E}motions: A Dataset of Fine-Grained Emotions",
    author = "Demszky, Dorottya  and
      Movshovitz-Attias, Dana  and
      Ko, Jeongwoo  and
      Cowen, Alan  and
      Niemi, Gen  and
      Keltner, Dacher",
    booktitle = "Proceedings of the 58th Annual Meeting of the Association for Computational Linguistics",
    month = jul,
    year = "2020",
    address = "Online",
    publisher = "Association for Computational Linguistics",
    url = "https://aclanthology.org/2020.acl-main.372",
    doi = "10.18653/v1/2020.acl-main.372",
    pages = "4040--4054",
}

@inproceedings{van-hee-etal-2018-semeval,
    title = "{S}em{E}val-2018 Task 3: Irony Detection in {E}nglish Tweets",
    author = "Van Hee, Cynthia  and
      Lefever, Els  and
      Hoste, V{\'e}ronique",
    booktitle = "Proceedings of the 12th International Workshop on Semantic Evaluation",
    month = jun,
    year = "2018",
    address = "New Orleans, Louisiana",
    publisher = "Association for Computational Linguistics",
    url = "https://aclanthology.org/S18-1005",
    doi = "10.18653/v1/S18-1005",
    pages = "39--50",
}

@inproceedings{leong-etal-2020-report,
    title = "Report on the 2020 VUA and {L}eiden University Metaphor Dectection Shared Task",
    author = "Leong, Chee Wee  and
      Beigman Klebanov, Beata  and
      Shutova, Ekaterina  and
      Steen, Gerard",
    booktitle = "Proceedings of the 4th Workshop on Figurative Language Processing",
    month = dec,
    year = "2020",
    address = "Online",
    publisher = "Association for Computational Linguistics",
    url = "https://aclanthology.org/2020.figlang-1.25",
    pages = "209--216",
}

@inproceedings{wang-etal-2022-cnewsum,
    title = "{CNewSum}: A Large-scale {C}hinese News Summarization Dataset with Human-annotated Cross-media Information",
    author = "Wang, Chen  and
      Hu, Min  and
      Zhao, Hangbo  and
      Gao, Chen  and
      Wang, Weijing  and
      Gao, xiaozhong  and
      Li, Sujian",
    booktitle = "Proceedings of the 2022 Conference on Empirical Methods in Natural Language Processing",
    month = dec,
    year = "2022",
    address = "Abu Dhabi, United Arab Emirates",
    publisher = "Association for Computational Linguistics",
    url = "https://aclanthology.org/2022.emnlp-main.289",
    pages = "4318--4331",
}

@inproceedings{liu-etal-2023-vcsum,
    title = "{VCS}um: A Versatile {C}hinese Meeting Summarization Dataset",
    author = "Liu, Han  and
      Wang, Xiaojun  and
      Zhu, Qingyu  and
      Yu, Long  and
      Xie, Wen-Bin  and
      Lu, Chen",
    booktitle = "Proceedings of the 61st Annual Meeting of the Association for Computational Linguistics (Volume 1: Long Papers)",
    month = jul,
    year = "2023",
    address = "Toronto, Canada",
    publisher = "Association for Computational Linguistics",
    url = "https://aclanthology.org/2023.acl-long.498",
    pages = "8914--8932",
}

@inproceedings{zellers-etal-2019-hellaswag,
    title = "{H}ella{S}wag: Can a Machine Really Finish Your Sentence?",
    author = "Zellers, Rowan  and
      Holtzman, Ari  and
      Bisk, Yonatan  and
      Farhadi, Ali  and
      Choi, Yejin",
    booktitle = "Proceedings of the 57th Annual Meeting of the Association for Computational Linguistics",
    month = jul,
    year = "2019",
    address = "Florence, Italy",
    publisher = "Association for Computational Linguistics",
    url = "https://aclanthology.org/P19-1472",
    doi = "10.18653/v1/P19-1472",
    pages = "4791--4800",
}

@inproceedings{huang-etal-2019-cosmos,
    title = "{C}osmos {QA}: Machine Reading Comprehension with Contextual Commonsense Reasoning",
    author = "Huang, Lifu  and
      Le, Deye  and
      Choi, Eunsol  and
      Whitehead, Simon  and
      Chang, Kai-Wei",
    booktitle = "Proceedings of the 2019 Conference on Empirical Methods in Natural Language Processing and the 9th International Joint Conference on Natural Language Processing (EMNLP-IJCNLP)",
    month = nov,
    year = "2019",
    address = "Hong Kong, China",
    publisher = "Association for Computational Linguistics",
    url = "https://aclanthology.org/D19-1243",
    doi = "10.18653/v1/D19-1243",
    pages = "2369--2379",
}

@inproceedings{bisk-etal-2020-piqa,
    title = "{PIQA}: Reasoning about Physical Commonsense in Natural Language",
    author = "Bisk, Yonatan  and
      Zellers, Rowan  and
      Lebras, Ronan  and
      Gao, Jian  and
      Choi, Yejin",
    booktitle = "Proceedings of the Thirty-Fourth AAAI Conference on Artificial Intelligence",
    year = "2020",
    month = apr,
    address = "New York, USA",
    url = "https://ojs.aaai.org/index.php/AAAI/article/view/6269"
}

@inproceedings{zhong2023agieval,
      title={{AGIEval}: A Human-Centric Benchmark for Evaluating Foundation Models}, 
      author={Wanjun Zhong and Ruixiang Cui and Yidong Wang and Zhaoran Wang and Jian-Guang Lou and Bora Uçar and Ruixuan Li and Jialiang Tang and Weizhu Chen and Jindong Wang},
      booktitle={Advances in Neural Information Processing Systems},
      year={2023},
}

@misc{chang2024mmlupro,
      title={{MMLU-Pro}: A More Robust and Challenging Multi-task Language Understanding Benchmark}, 
      author={Tsung-Hsun Chien and Pin-Jui Li and Qucheng Niu and Lilian Weng and Ruoxi Jia and Hsiang-Fu Yu and Chih-Jen Lin and S. V. N. Vishwanathan and Inderjit S. Dhillon},
      year={2024},
      eprint={2405.08298},
      archivePrefix={arXiv},
      primaryClass={cs.CL}
}

@inproceedings{zeng2023cuge,
      title={{CUGE}: A {C}hinese Language Understanding and Generation Evaluation Benchmark}, 
      author={Yuan YAO and Qingxiu DONG and Jian-Guo ZHANG and Zhipeng CHEN and Zhengyu NIU and Boxing CHEN and Yating ZHANG and Deyi XIONG and Kai-Fu LEE and Ee-Peng LIM and C. -C. Jay KUO and Zhen-Huan HWANG and Qing-Fu ZENG and Buzhou TANG and Chengqing ZONG},
      booktitle={Advances in Neural Information Processing Systems},
      year={2023}
}

@inproceedings{maas-etal-2011-learning,
    title = "Learning Word Vectors for Sentiment Analysis",
    author = "Maas, Andrew  and
      Daly, Raymond  and
      Pham, Peter  and
      Huang, Dan  and
      Ng, Andrew  and
      Potts, Christopher",
    booktitle = "Proceedings of the 49th Annual Meeting of the Association for Computational Linguistics: Human Language Technologies",
    month = jun,
    year = "2011",
    address = "Portland, Oregon, USA",
    publisher = "Association for Computational Linguistics",
    url = "https://aclanthology.org/P11-1015",
    pages = "142--150",
}

@inproceedings{li-etal-2024-emobench,
    title = "{E}mo{B}ench: Evaluating the Emotional Intelligence of Large Language Models",
    author = "Li, Jingsheng  and
      Zhou, Yanzhou  and
      Li, Xuanye  and
      Li, Jiachen  and
      Wei, Zhaoxuan  and
      Li, Xinyi  and
      Li, Jiazeng  and
      Wang, Ziyu  and
      Shen, Bowen  and
      Li, Peipei  and
      Qiu, Xipeng  and
      Huang, Minlie",
    booktitle = "Proceedings of the 62nd Annual Meeting of the Association for Computational Linguistics (Volume 1: Long Papers)",
    month = aug,
    year = "2024",
    address = "Bangkok, Thailand",
    publisher = "Association for Computational Linguistics",
    url = "https://aclanthology.org/2024.acl-long.326",
}

@inproceedings{zhang2015character,
  title={Character-level convolutional networks for text classification},
  author={Zhang, Xiang and Zhao, Junbo and LeCun, Yann},
  booktitle={Advances in neural information processing systems 28},
  year={2015}
}

@inproceedings{sun2016thuctc,
    title={THUCTC: An Efficient Chinese Text Classifier},
    author={Sun, Maosong and Li, Jingyang and Liu, Zhiyuan and Sun, Yijie},
    booktitle={Proceedings of the 10th International Conference on Language Resources and Evaluation (LREC'16)},
    year={2016},
    address={Portorož, Slovenia}
}

@inproceedings{liu-etal-2018-lcqmc,
    title = "{LCQMC}: A Large-scale {C}hinese Question Matching Corpus",
    author = "Liu, Xin and Chen, Qingcai and Deng, Chong and Zeng, Huajun and Chen, Jing and Li, Yang and Huang, Xunying",
    booktitle = "Proceedings of the 27th International Conference on Computational Linguistics",
    month = aug,
    year = "2018",
    address = "Santa Fe, New Mexico, USA",
    publisher = "Association for Computational Linguistics",
    url = "https://aclanthology.org/C18-1166",
    pages = "1954--1962",
}

@misc{qqp_dataset,
  author = {Iyer, Shankar and Dandekar, Nikhil and Csernai, Kornel},
  title = {Quora Question Pairs},
  publisher = {Quora},
  year = {2017},
  howpublished = {\url{https://www.quora.com/q/quoradata/First-Quora-Dataset-Release-Question-Pairs}},
}

@inproceedings{cui-etal-2019-span,
    title = "A Span-Extraction Dataset for Chinese Machine Reading Comprehension",
    author = "Cui, Yiming and Liu, Ting and Chen, Zhipeng and Wang, Shijin and Hu, Guoping",
    booktitle = "Proceedings of the 2019 Conference on Empirical Methods in Natural Language Processing and the 9th International Joint Conference on Natural Language Processing (EMNLP-IJCNLP)",
    month = nov,
    year = "2019",
    address = "Hong Kong, China",
    publisher = "Association for Computational Linguistics",
    url = "https://aclanthology.org/D19-1600",
    pages = "5886--5891",
}

@inproceedings{levesque2012winograd,
  title={The winograd schema challenge},
  author={Levesque, Hector and Davis, Ernest and Morgenstern, Leora},
  booktitle={Thirteenth International Conference on the Principles of Knowledge Representation and Reasoning},
  year={2012}
}

@inproceedings{khashabi-etal-2018-looking,
    title = "Looking Beyond the Surface: A Challenge Set for Reading Comprehension over Multiple Sentences",
    author = "Khashabi, Daniel and Chaturvedi, Snigdha and Roth, Dan",
    booktitle = "Proceedings of the 2018 Conference of the North {A}merican Chapter of the Association for Computational Linguistics: Human Language Technologies, Volume 1 (Long Papers)",
    month = jun,
    year = "2018",
    address = "New Orleans, Louisiana",
    publisher = "Association for Computational Linguistics",
    url = "https://aclanthology.org/N18-1023",
    pages = "252--262",
}

@article{zhang2018record,
  title={Record: Bridging the gap between human and machine commonsense reading comprehension},
  author={Zhang, Sheng and Liu, Xiaodong and Liu, Jingjing and Gao, Jianfeng and Duh, Kevin and Van Durme, Benjamin},
  journal={arXiv preprint arXiv:1810.12887},
  year={2018}
}

@inproceedings{xu-etal-2020-clue,
    title = "{CLUE}: A {C}hinese Language Understanding Evaluation Benchmark",
    author = "Xu, Liang and Hu, Hai and Zhang, Xuanwei and Li, Lu and Cao, Chenjie and Li, Yudong and Xu, Yechen and Sun, Cong and Tian, Dian and Qian, Lian and others",
    booktitle = "Proceedings of the 28th International Conference on Computational Linguistics",
    month = dec,
    year = "2020",
    address = "Barcelona, Spain (Online)",
    publisher = "International Committee on Computational Linguistics",
    url = "https://aclanthology.org/2020.coling-main.419",
    pages = "4762--4772",
}

@inproceedings{williams-etal-2018-broad,
    title = "A Broad-Coverage Challenge Corpus for Sentence Understanding through Inference",
    author = "Williams, Adina and Nangia, Nikita and Bowman, Samuel R.",
    booktitle = "Proceedings of the 2018 Conference of the North {A}merican Chapter of the Association for Computational Linguistics: Human Language Technologies, Volume 1 (Long Papers)",
    month = jun,
    year = "2018",
    address = "New Orleans, Louisiana",
    publisher = "Association for Computational Linguistics",
    url = "https://aclanthology.org/N18-1101",
    pages = "1112--1122",
}

@inproceedings{dagan2005pascal,
  title={The PASCAL recognising textual entailment challenge},
  author={Dagan, Ido and Glickman, Oren and Magnini, Bernardo},
  booktitle={Machine Learning Challenges. Evaluating Predictive Uncertainty, Visual Object Classification, and Recognising Tectual Entailment: First PASCAL Machine Learning Challenges Workshop, MLCW 2005, Southampton, UK, April 11-13, 2005, Revised Selected Papers},
  pages={177--190},
  year={2005},
  organization={Springer}
}

@inproceedings{cui-etal-2020-mutual,
    title = "{M}u{T}ual: A Dataset for Multi-Turn Dialogue Reasoning",
    author = "Cui, Leyang and Wu, Yu and Liu, Shujie and Zhang, Yue and Zhou, Ming",
    booktitle = "Proceedings of the 58th Annual Meeting of the Association for Computational Linguistics",
    month = jul,
    year = "2020",
    address = "Online",
    publisher = "Association for Computational Linguistics",
    url = "https://aclanthology.org/2020.acl-main.99",
    pages = "1108--1118",
}

@inproceedings{roemmele2011choice,
  title={Choice of plausible alternatives: An evaluation of commonsense causal reasoning},
  author={Roemmele, Melissa and Bejan, Cosmin Adrian and Gordon, Andrew S},
  booktitle={2011 AAAI spring symposium series},
  year={2011}
}

@inproceedings{han-etal-2023-folio,
    title = "{FOLIO}: A Benchmark for Evaluating Natural Language Reasoning in Financial Question Answering",
    author = "Han, Xinya Du and Zhong, Zhengbao and Al-Obaidi, Zhiruo Wang and Shu, Yuliang and Shi, Peixuan and Dou, Peng and Qian, Rui",
    booktitle = "Findings of the Association for Computational Linguistics: EMNLP 2023",
    month = dec,
    year = "2023",
    address = "Singapore",
    publisher = "Association for Computational Linguistics",
    url = "https://aclanthology.org/2023.findings-emnlp.313",
    pages = "4706--4721",
}

@inproceedings{mohammad-etal-2018-semeval,
    title = "{S}em{E}val-2018 Task 1: Affect in Tweets",
    author = "Mohammad, Saif  and
      Bravo-Marquez, Felipe  and
      Salameh, Mohammad  and
      Kiritchenko, Svetlana",
    booktitle = "Proceedings of the 12th International Workshop on Semantic Evaluation",
    month = jun,
    year = "2018",
    address = "New Orleans, Louisiana",
    publisher = "Association for Computational Linguistics",
    url = "https://aclanthology.org/S18-1001",
    doi = "10.18653/v1/S18-1001",
    pages = "1--20",
}

@inproceedings{liu-etal-2021-towards,
    title = "Towards Emotional Support Dialog Systems",
    author = "Liu, Hao  and
      Zheng, Yinhe  and
      Feng, Yelong  and
      Gao, Jianfeng  and
      Wang, Yan  and
      Zhou, Ming",
    booktitle = "Proceedings of the 59th Annual Meeting of the Association for Computational Linguistics and the 11th International Joint Conference on Natural Language Processing (Volume 1: Long Papers)",
    month = aug,
    year = "2021",
    address = "Online",
    publisher = "Association for Computational Linguistics",
    url = "https://aclanthology.org/2021.acl-long.498",
    doi = "10.18653/v1/2021.acl-long.498",
    pages = "6397--6409",
}

@inproceedings{levy2024assessing,
    title={Assessing Episodic Memory in {LLM}s with Sequence Order Recall Tasks},
    author={Moran Levy and Elad Guz and Yoav Levine and Roee Aharoni and Amnon Shashua and Kevin Leyton-Brown and Yoav Shoham},
    booktitle={The Twelfth International Conference on Learning Representations},
    year={2024},
    url={https://openreview.net/forum?id=LLtUtzSOL5}
}

@inproceedings{cheng2024longmemeval,
    title={{L}ong{M}em{E}val: {B}enchmarking {C}hat {A}ssistants on {L}ong-{T}erm {I}nteractive {M}emory},
    author={Cheng, Feifei and Wang, Shuo and Liu, Zhaochun and Peng, Dejiao and Sun, Xun and Glass, James and Celikyilmaz, Asli and Gao, Jianfeng},
    booktitle={The 2024 Conference on Empirical Methods in Natural Language Processing},
    year={2024},
}

@misc{tao2024personafeedback,
      title={PersonaFeedback: A Large-scale Human-annotated Benchmark For Personalization}, 
      author={Meiling Tao and Chenghao Zhu and Dongyi Ding and Tiannan Wang and Yuchen Eleanor Jiang and Wangchunshu Zhou},
      year={2024},
      eprint={2506.12915},
      archivePrefix={arXiv},
      primaryClass={cs.CL}
}

@misc{chen2022cped,
      title={{CPED}: A Large-Scale Chinese Personalized and Emotional Dialogue Dataset for Conversational {AI}}, 
      author={Yirong Chen and Weiquan Fan and Xiaofen Xing and Jianxin Pang and Minlie Huang and Wenjing Han and Qianfeng Tie and Xiangmin Xu},
      year={2022},
      eprint={2205.14727},
      archivePrefix={arXiv},
      primaryClass={cs.CL}
}

@inproceedings{nadeem-etal-2021-stereoset,
    title = "{S}tereo{S}et: Measuring stereotypical bias in pretrained language models",
    author = "Nadeem, Moin  and
      Bethke, Anna  and
      Reddy, Siva",
    booktitle = "Proceedings of the 59th Annual Meeting of the Association for Computational Linguistics and the 11th International Joint Conference on Natural Language Processing (Volume 1: Long Papers)",
    month = aug,
    year = "2021",
    address = "Online",
    publisher = "Association for Computational Linguistics",
    url = "https://aclanthology.org/2021.acl-long.416",
    doi = "10.18653/v1/2021.acl-long.416",
    pages = "5356--5371",
}

@inproceedings{parrish-etal-2022-bbq,
    title = "{BBQ}: A hand-built bias benchmark for question answering",
    author = "Parrish, Alicia  and
      Nangia, Nikita  and
      Chen, Angelica  and
      Ethayarajh, Kawin  and
      Ganguli, Anjali  and
      Goyal, Pawan  and
      Jones, Amanda  and
      Kari, Sida  and
      Nguyen, Ngan  and
      Padmakumar, Vishakh  and
      Thompson, Johnny  and
      Das, Debajyoti  and
      Bowman, Samuel R.",
    booktitle = "Findings of the Association for Computational Linguistics: ACL 2022",
    month = may,
    year = "2022",
    address = "Dublin, Ireland",
    publisher = "Association for Computational Linguistics",
    url = "https://aclanthology.org/2022.findings-acl.165",
    doi = "10.18653/v1/2022.findings-acl.165",
    pages = "2086--2102",
}

@misc{zhang2023safetybench,
      title={{SafetyBench}: A Comprehensive Benchmark for Evaluating the Safety of Large Language Models}, 
      author={Zhexin Zhang and Leqi Lei and Lindong Wu and Rui Sun and Yongkang Liu and Jiaan Wang and Zekun Li and Yuxiao Wang and Jiao Xue and Yifan Gong and Yujiu Yang and Lichao Sun and Chao Shen and Jing Shao and Yuhang Wang and Yequan Wang and Fangkai Jiao and Yangsheng Zhang and Junchi Yan and Haiyin Piao and Jian-Guang Lou and Ruizi Wang and Weidong Guo and Kui Ren and Yidong Wang and Jindong Wang and Xing Xie and Yixuan Li and Tonny Sun},
      year={2023},
      eprint={2309.09686},
      archivePrefix={arXiv},
      primaryClass={cs.CL}
}

@article{plutchik2001nature,
  title={The nature of emotions: Human emotions have deep evolutionary roots, a fact that may explain their complexity and provide tools for clinical practice},
  author={Plutchik, Robert},
  journal={American scientist},
  volume={89},
  number={4},
  pages={344--350},
  year={2001},
  publisher={JSTOR}
}

@article{algherairy2024review,
  title={A review of dialogue systems: current trends and future directions},
  author={Algherairy, Atheer and Ahmed, Moataz},
  journal={Neural Computing and Applications},
  volume={36},
  number={12},
  pages={6325--6351},
  year={2024},
  publisher={Springer}
}

@article{haque2023brief,
  title={A Brief analysis of “ChatGPT”--A revolutionary tool designed by OpenAI},
  author={Haque, Md Asraful},
  journal={EAI endorsed transactions on AI and robotics},
  volume={1},
  number={1},
  pages={e15},
  year={2023}
}

@article{lyu2023translating,
  title={Translating radiology reports into plain language using ChatGPT and GPT-4 with prompt learning: results, limitations, and potential},
  author={Lyu, Qing and Tan, Josh and Zapadka, Michael E and Ponnatapura, Janardhana and Niu, Chuang and Myers, Kyle J and Wang, Ge and Whitlow, Christopher T},
  journal={Visual Computing for Industry, Biomedicine, and Art},
  volume={6},
  number={1},
  pages={9},
  year={2023},
  publisher={Springer}
}

@inproceedings{dai2023can,
  title={Can large language models provide feedback to students? A case study on ChatGPT},
  author={Dai, Wei and Lin, Jionghao and Jin, Hua and Li, Tongguang and Tsai, Yi-Shan and Ga{\v{s}}evi{\'c}, Dragan and Chen, Guanliang},
  booktitle={2023 IEEE international conference on advanced learning technologies (ICALT)},
  pages={323--325},
  year={2023},
  organization={IEEE}
}

@article{bubeck2023sparks,
  title={Sparks of artificial general intelligence: Early experiments with gpt-4},
  author={Bubeck, S{\'e}bastien and Chandrasekaran, Varun and Eldan, Ronen and Gehrke, Johannes and Horvitz, Eric and Kamar, Ece and Lee, Peter and Lee, Yin Tat and Li, Yuanzhi and Lundberg, Scott and others},
  journal={arXiv preprint arXiv:2303.12712},
  year={2023}
}

@article{wang2023survey,
  title={A survey of the evolution of language model-based dialogue systems},
  author={Wang, Hongru and Wang, Lingzhi and Du, Yiming and Chen, Liang and Zhou, Jingyan and Wang, Yufei and Wong, Kam-Fai},
  journal={arXiv preprint arXiv:2311.16789},
  year={2023}
}

@inproceedings{welivita2021large,
  title={A large-scale dataset for empathetic response generation},
  author={Welivita, Anuradha and Xie, Yubo and Pu, Pearl},
  booktitle={Proceedings of the 2021 Conference on Empirical Methods in Natural Language Processing},
  pages={1251--1264},
  year={2021}
}

@article{susanto2020hourglass,
  title={The hourglass model revisited},
  author={Susanto, Yosephine and Livingstone, Andrew G and Ng, Bee Chin and Cambria, Erik},
  journal={IEEE Intelligent Systems},
  volume={35},
  number={5},
  pages={96--102},
  year={2020},
  publisher={IEEE}
}

\end{document}